\documentclass{bmvc2k}

\usepackage{epsfig}
\usepackage{graphicx}
\usepackage{amsmath}
\usepackage{amssymb}
\usepackage{wrapfig}
\usepackage{floatrow}
\usepackage{array}
\usepackage{pythonhighlight}
\usepackage{comment}
\usepackage{nicefrac}

\usepackage{inconsolata}

\usepackage{algorithm}
\usepackage{algpseudocode}

\usepackage{multirow}
\usepackage{rotating}
\usepackage{booktabs}

\usepackage{enumitem}
\usepackage[olditem,oldenum]{paralist}

\usepackage{alltt}
\usepackage{listings}

\newcommand{\csection}[1]{
    \vspace{-0.1in}
    \section{#1}
    \vspace{-0.06in}
}

\newcommand{\csubsection}[1]{
    \vspace{-0.1in}
    \subsection{#1}
    \vspace{-0.05in}
}

\usepackage{url}
\usepackage{xspace}
\usepackage{comment}
\usepackage{color}
\usepackage{xcolor}
\usepackage{framed}
\usepackage{fancybox}
\usepackage{cuted}
\usepackage{paralist}

\usepackage{footnote}
\makesavenoteenv{tabular}

\definecolor{citecolor}{HTML}{2779af}
\definecolor{linkcolor}{HTML}{c0392b}
\definecolor{darkgreen}{HTML}{27ae60}

\PassOptionsToPackage{pagebackref=false,breaklinks=true,colorlinks,bookmarks=false,citecolor=citecolor,linkcolor=linkcolor}{hyperref}

\title{Realistic PointGoal Navigation via Auxiliary Losses and Information Bottleneck}

\addauthor{Guillermo N. Grande}{nicogrande@gatech.edu}{1}
\addauthor{Dhruv Batra}{dbatra@gatech.edu}{12}
\addauthor{Erik Wijmans}{etw@gatech.edu}{12}

\addinstitution{
 Georgia Institute of Technology \\
 Atlanta, GA
}
\addinstitution{
 Facebook AI Research \\
 Menlo Park, CA
}

\runninghead{Grande, Batra, and Wijmans}{Realistic PointGoal Navigation}

\def\etal{\emph{et al}\bmvaOneDot}

\newcommand{\myquote}[1]{\emph{`#1'}}
\newcommand{\myapprox}{{\raise.17ex\hbox{$\scriptstyle\sim$}}}
\newcommand{\xhdr}[1]{\vspace{0pt}\noindent\textbf{#1}\xspace}

\newcommand{\reffig}[1]{Fig.~\ref{#1}}
\newcommand{\refsec}[1]{Sec.~\ref{#1}}
\newcommand{\reftab}[1]{Tab.~\ref{#1}}

\makeatletter
\newcommand\footnoteref[1]{\protected@xdef\@thefnmark{\ref{#1}}\@footnotemark}
\makeatother

\newcommand{\rgbd}{\texttt{RGB-D}\xspace}

\newcommand{\depth}{\texttt{Depth}\xspace}

\newcommand{\pointnav}{\texttt{PointGoalNav}\xspace}
\newcommand{\pointnavfull}{PointGoal navigation\xspace}
\newcommand{\compassgps}{GPS+Compass\xspace}
\newcommand{\gpscompass}{\compassgps}

\begin{document}

\maketitle

\begin{abstract}

We propose a novel architecture and training paradigm for training realistic \pointnavfull ~-- navigating to a target coordinate in an unseen environment under actuation and sensor noise \emph{without} access to ground-truth localization.  Specifically, we find that the primary challenge under this setting is learning localization -- when stripped of idealized localization, agents fail to stop precisely at the goal despite reliably making progress towards it. To address this we introduce a set of auxiliary losses to help the agent learn localization.  Further, we explore the idea of treating the precise location of the agent as privileged information -- it is unavailable during test time, however, it is available during training time in simulation. We grant the agent \emph{restricted} access to ground-truth localization readings during training via an information bottleneck. Under this setting, the agent incurs a penalty for using this privileged information, encouraging the agent to only leverage this information when it is crucial to learning. This enables the agent to first learn navigation and then learn localization instead of conflating these two objectives in training.  We evaluate our proposed method both in a semi-idealized (noiseless simulation \emph{without} Compass+GPS) and realistic (addition of noisy simulation) settings. Specifically, our method outperforms existing baselines on the semi-idealized setting by 18\%/21\% SPL/Success and by 15\%/20\% SPL in the realistic setting. Our improved Success and SPL metrics indicate our agent's improved ability to accurately self-localize while maintaining a strong navigation policy. 
Our implementation can be found at \href{https://github.com/NicoGrande/habitat-pointnav-via-ib}{https://github.com/NicoGrande/habitat-pointnav-via-ib}.

\end{abstract}

\csection{Introduction}

There has been substantial progress made in training agents to navigate 3D simulated environments from egocentric perception via reinforcement learning (RL) ~\cite{ddppo,habitat19iccv,anderson2018vision,embodiedqa,wijmans2019embodied,gordon2018iqa}. These embodied agents perceive their environment using a predefined sensor suite to condition their actions as they perform a given task. One foundational task is \pointnavfull~\cite{anderson2018evaluation} wherein an embodied agent is tasked with navigating to a coordinate specified relative to its position at the start of an episode. The agent must do so in a previously unseen environment without access to a map. To succeed, the agent must stop within a fixed distance of the target coordinate. 
Recent work~\cite{ddppo} has shown that this task can be performed nearly perfectly in simulation under the assumption that agents are equipped with ground-truth localization information (\gpscompass), noiseless depth sensing, and idealized actuators.  

\begin{wrapfigure}{R}{0.5\textwidth}
\centering
\includegraphics[width=0.7\textwidth]{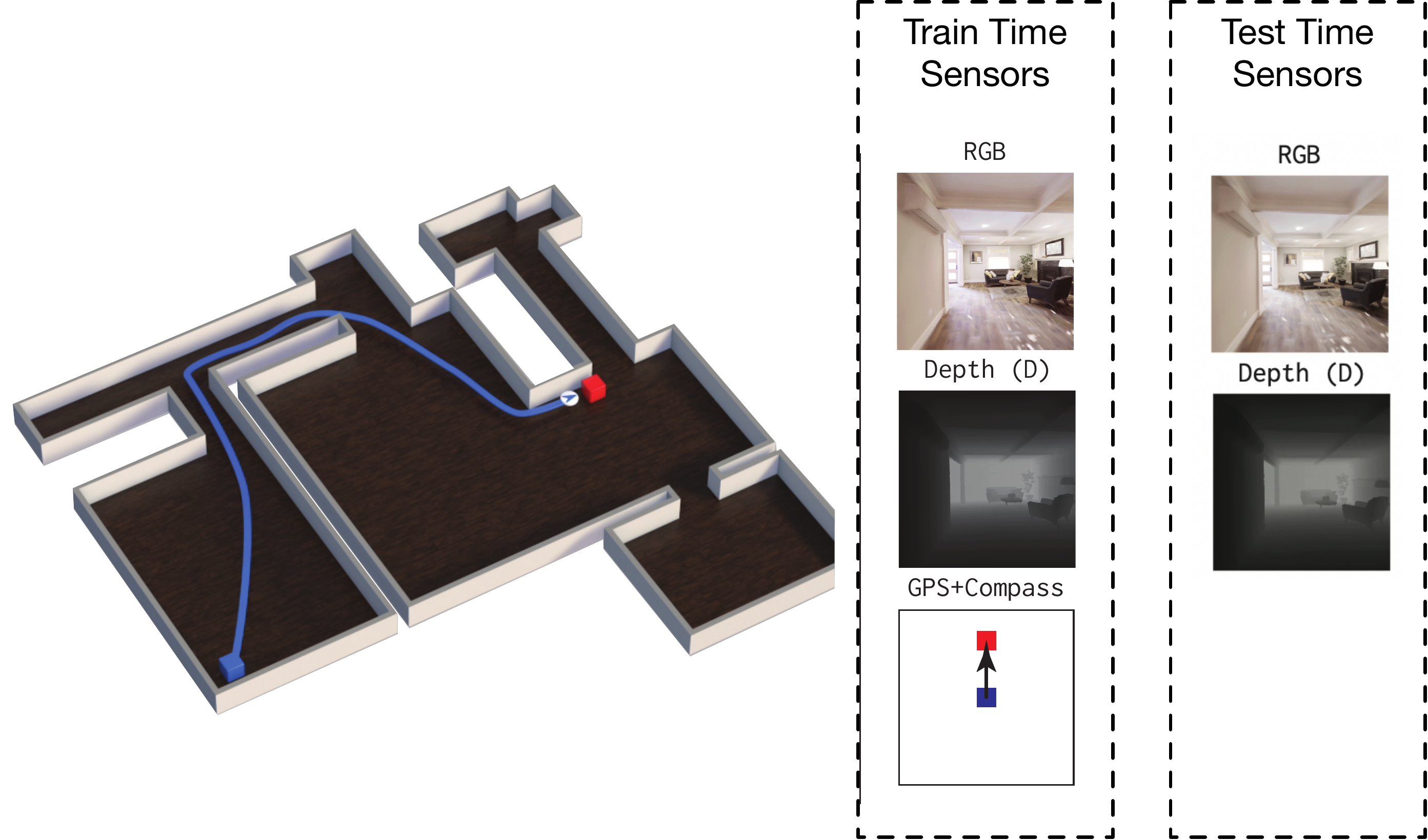}
\caption{\textbf{\pointnavfull with privileged information training}. 
} \label{fig:pointnav}
\end{wrapfigure}

\pointnavfull is far from solved when these assumptions no longer hold. It is unrealistic to rely on idealized localization information in previously unseen environments given current technology~\cite{localization}. Similarly, it is unrealistic to assume perfect sensing (real \depth cameras have noise~\cite{camera_noise}) and fully deterministic actuations (real robots do not arrive at precisely the commanded location). Removing idealized localization, adding noise to cameras, and adding stochasticity to actuations much more closely resembles reality. Effectively performing \pointnavfull under these more realistic conditions is the subject of 2020 Habitat Challenge and the focus of this work.  

Wijmans \etal~\cite{ddppo} showed that with ground truth localization, \gpscompass, \pointnavfull can be  `solved' via end-to-end reinforcement learning. Thus, it is natural to ask -- \myquote{How well do model-free Reinforcement Learning techniques translate from idealized to realistic \pointnavfull?} We find evidence suggesting that the direct transfer fails catastrophically with the key problem being that the agent does not learn to self-localize.

In an end-to-end learning setting, the agent must not only learn to navigate effectively given only visual inputs, but also learn to track itself. Through this work we find that these tasks are highly interdependent: navigating effectively towards the target coordinate is facilitated by good tracking, while tracking must be learned with respect to the current policy. Specifically, we find that as tracking improves, this enables learning a better navigation policy thereby \emph{changing the distribution of inputs} and degrading tracking performance. As such, conflating these two into a single objective leads to sub-optimal results, when attempting to jointly learn tracking and navigation. %

Given this key observation, we motivate our primary idea -- we propose serializing these two objectives by treating idealized \gpscompass as privileged information available during training (in simulation) but absent at evaluation time. This privileged information allows the agent to focus on learning navigation and then, once it has a reasonable navigation policy, focus on learning tracking. We structure our proposed learning paradigm via an information bottleneck~\cite{tishby99information} -- a method to limit the amount of information a latent variable can encode about its inputs. The information bottleneck ensures that the agent will leverage privileged information only when it is crucial to progressing in a \pointnavfull episode, preventing the agent from becoming reliant on this information at evaluation time.  We show in [\reffig{fig:aux_losses}] of the supplementary section that this approach leads to lower self-localization error overall, supporting our hypothesis that conflating navigation and self-localization is harmful to self-localization performance. Additionally, we find using a visual-encoder pretrained on \pointnavfull \emph{with} \gpscompass facilitates faster learning. 
we introduce a set of auxiliary losses to encourage the embodied agent to learn objectives relevant to successful \pointnavfull without \gpscompass. These include estimating the agent's change in pose (ego-motion), and estimating \gpscompass sensor readings (tracking).

Experimentally, we find that in a noiseless semi-idealized setting (idealized sensing and actuation \emph{without} idealized localization) our agent is able to learn high performing policies with high sample efficiency \emph{without} \gpscompass at evaluation time, achieving 69\%/75\% SPL/Success, an improvement of 18\%/21\% SPL/Success over existing baselines. In the noisy 2020 Habitat Challenge setting our model achieves 20\%/26\% SPL/Success, outperforming existing baselines by 15\%/20\% SPL. Through a set of ablation studies we find that the proposed auxiliary losses are fundamental to the agent's ability to self-localize, while the information bottleneck yields modest gains in sample efficiency and performance.

\csection{Related Work}

Goyal \etal \cite{goyal2019infobot} initially proposed the idea of leveraging an information bottleneck for reinforcement learning to improve sample efficiency when learning with sparse rewards signals.   In this work, we use an information bottleneck to give the agent constrained access to ground truth \gpscompass thereby enabling the agent to learn with a different sensor suite than is available during evaluation.

Analogously to our approach, Chen \etal \cite{chen2019learning} train a sensorimotor agent which is given access to privileged information at training time.  Unlike our approach however, Chen \etal then learn a second agent which attempts to imitate that privileged agent. We forego imitation learning in our approach given the issues of covariate shift associated with behavior cloning when exploring previously unseen environments~\cite{DBLP:journals/corr/LaskeyLHLMFG17}. Pinto \etal~\cite{pinto2017asymmetric} also use privileged information in reinforcement learning, specifically they use a different sensor-suite for the value network and policy network, giving privileged information to the value network.  We condition both our policy and value network on privileged information at training time. Kamienny \etal \cite{PI-Dropout} grant privileged information to a grid-world agent via privileged information dropout, a method inspired by information bottleneck theory~\cite{tishby99information}. 

\xhdr{Visual Navigation.}  There has been significant interest in visual navigation recently, with works developing methods for \pointnavfull with ground truth \gpscompass~\cite{chaplot2020learning,ramakrishnan2020occupancy,ddppo, bansal2019combining}, ObjectGoal navigation~\cite{objectnav,chaplot2020object, DBLP:journals/corr/GuptaDLSM17}, RoomGoal navigation~\cite{narasimhan2020seeing}, ImageGoal navigation~\cite{zhu2017target} and more. We focus on \pointnavfull under realistic settings \emph{without} access to ground truth \gpscompass, as advancements in this setting transfer well to reality~\cite{Kadian_2020}.

\xhdr{Realistic \pointnavfull methods.}
\pointnavfull under realistic conditions without \gpscompass has also been approached in other works. 
Datta \etal \cite{datta2020integrating} used an agent trained with ground truth GPS+Compass to collect an egomotion dataset used to train an odometery model to predict changes in pose given consecutive \rgbd observations and then integrate these deltas. 
In contrast, our approach does this with auxiliary losses and serialization is done via an optimization objective as opposed to manually imposed.

Ramakrishnan \etal \cite{ramakrishnan2020occupancy} approach this task by using egocentric \rgbd observations to anticipate occupancy of obstacles in a given scene. 
While Ramakrishnan \etal show that leveraging an internal map is effective for realistic PointGoal navigation, we use a simple policy network as we find end-to-end trainable policy methods to be less computationally expensive at evaluation time than map based methods [\reftab{tab:habitat-challenge-2020-results}, Col 4], a key consideration when deploying models in reality.
 \csection{Preliminaries: Task, Agent, and Training}
\label{sec:background}

\xhdr{Task.} In \pointnavfull~\cite{anderson2018evaluation} an agent is initialized at an arbitrary location in a unseen 3D environment. The agent is then prompted to navigate to a target specified relative to the agent's starting coordinate (i.e. \myquote{Go 5 meters north, 3 meters west relative to start} specified as the point (3, 5)). 

\xhdr{Agent Sensor Suite.} In this work we use a different sensor suite at training time and testing time.  At training time, the sensor suite of the agent at training is the following: \depth sensor (providing ego-centric depth-maps), GPS sensor (providing the agent its position relative to start), and Compass sensor (providing the agent access to its orientation relative to start).
At evaluation time, only the \depth sensor is available to the agent. When working under the realistic (Habitat Challenge 2020) setting, the Habitat simulator applies RedWood noise to the \depth observations. 

\xhdr{Agent Specifications.} The agent is modeled as a 0.88 meter tall cylinder with a radius of 0.18 meters as required by the Habitat Challenge specifications. The agents action set is composed of $\left\{\texttt{stop}, \texttt{move\_forward}, \texttt{turn\_right}, \texttt{turn\_left}\right\}$. The agent uses the \texttt{stop} action to signal that it believes it has reached the goal. \texttt{move\_forward}  moves the agent 0.25 meters forward, while \texttt{turn\_right} and \texttt{turn\_left} rotate the agent 30 degrees right and left, respectively. Furthermore, in the realistic setting, the actuation noise of our agent is modeled to simulate real world a LocoBot~\cite{10.1007/978-3-642-33515-0_45} as is standard in the 2020 Habitat Challenge. 

\xhdr{Evaluation.}
Performance is evaluated using four key metrics: Success, SPL~\cite{anderson2018evaluation}, Soft-SPL~\cite{datta2020integrating}, and Distance to Goal. Distance to Goal measures the agent's geodesic distance to the target when the \texttt{stop}  action was called. Success indicates whether the agent predicted \texttt{stop} within $R_\text{PG}$ of the goal, where $R_\text{PG} = 0.36$ meters in all of our experiments. SPL weighs the Success value by the length of the agent's path compared to the optimal path. Soft-SPL replaces the binary Success indicator with percent progress towards goal. 

\xhdr{PPO.}
Proximal Policy Optimization (PPO)~\cite{schulman2017ppo} is a class of policy-gradient reinforcement learning algorithm, which performs on-policy updates. Given a policy $\pi_{\theta}$ parametrized by $\theta$ as well as a set of trajectories or ``rollouts'' collected with $\pi_{\theta_{\text{old}}}$, PPO updates its policy parameters $\theta$ as by maximizing the following:
\begin{align}
    \mathcal{J}^\text{PPO}(\theta) = 
     \mathbb{E} \left[\min(r(\theta)\hat{A},\text{clip}(r(\theta), 1 - \epsilon, 1 + \epsilon) \hat{A}) \right]
\end{align}
where $\hat{A}$ is the estimated advantage and $r(\theta) = \nicefrac{\pi_{\theta} (a \mid o)}{ \pi_{\theta_{\text{old}}} (a \mid o)}$ is the ratio of action probabilities under the current policy $\pi_{\theta}$ and the trajectory collection policy $\pi_{\theta_\text{old}}$. 

\xhdr{Training.}  We use the AI Habitat platform~\cite{habitat19iccv} and scenes from the Gibson dataset~\cite{xia2018gibson} for training and evaluation.  We train our agent using Decentralized Distributed PPO (DD-PPO)~\cite{ddppo}, a distributed version of PPO~\cite{schulman2017ppo}.  We use a learning rate of $2.5\times10^{-4}$ with the Adam optimizer~\cite{kingma2015adam}.  We set $\gamma{=}0.99$ and GAE-$\lambda$~\cite{schulman2016high} to $0.95$.  We collect rollouts of length 96 from 6 environments running in parallel.  We perform 2 epochs of PPO with 2 mini-batches.  We use DD-PPO to train on 8 GPUs in parallel.

At each time-step, the agent is given a shaped reward.  Specifically, let $s_t$ be the current location, and $a_t$ be action taken, then the reward is
\begin{equation}
    r(s_t, a_t) = \begin{cases}
      \text{Success} \cdot 2.5 \cdot e^\frac{R_\text{PG} - \text{DistToGoal}}{R_\text{PG}} & \text{if } a_t = \texttt{stop} \\
    -\Delta_{\text{GeoDist}}(s_t, a_t) - \lambda & \text{otherwise}
    \end{cases}
\end{equation}
where $\Delta_{\text{GeoDist}}(s_t, a_t)$ is the change in geodesic distance to goal, $\lambda({=}0.01)$ is a slack penalty, and $R_\text{PG}$ is the success distance.  This reward introduces a novel additional exponential term that encourages the agent to stop closer to the center of the success zone as we find that otherwise the agent tends to stop along the perimeter, leading to failed episodes when tracking is slightly off. We refer to this exponential term as dynamic success reward (DSR).

 \csection{Methodology}
\label{sec:method}

\begin{figure*}
\centering
\includegraphics[width=0.9\textwidth]{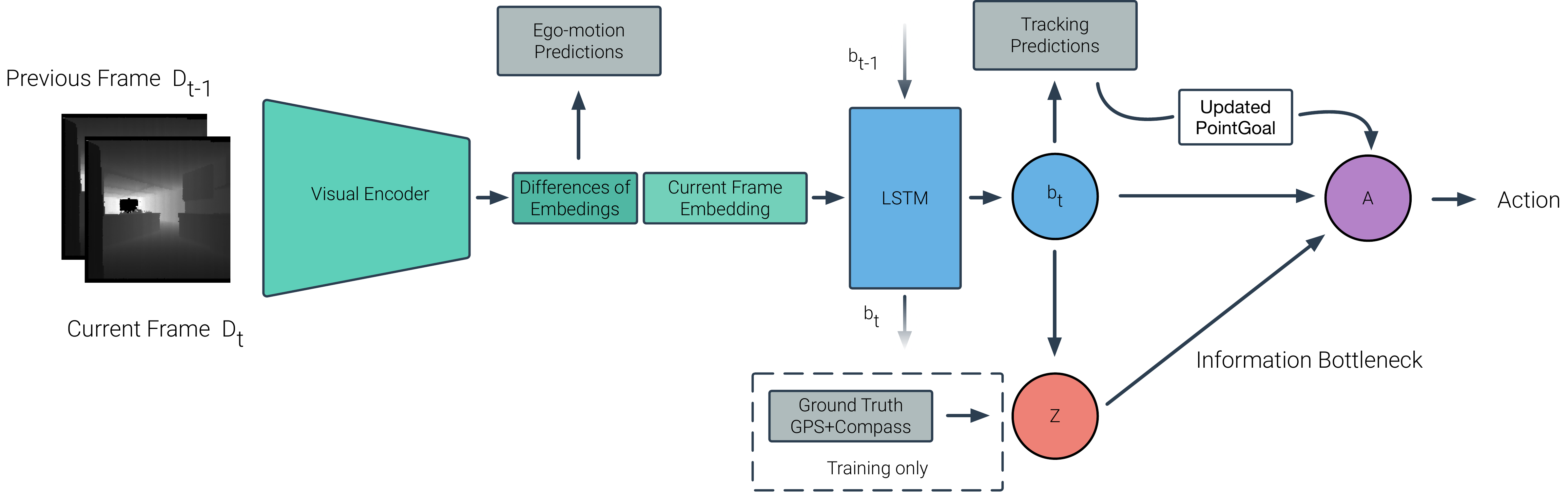}
\caption{\xhdr{Policy architecture with information bottleneck.} 
In our architecture, the current and previous observations are fed into a visual encoder. The embedding of the current frame and difference of embeddings between current and previous are then fed into a LSTM that updates the belief states $b_t$. The belief state is combined with the readings from the ground truth \gpscompass sensor available at training time and encoded into a latent variable $Z$. The amount of information $Z$ contains about \gpscompass is constrained by the information bottleneck. The belief state $b_t$ and $Z$ are then used to predict the action to take.  
}
\label{fig:model}
\end{figure*}
 
Our model takes consecutive depth observations as input and outputs the most likely action given the input observations and the agent's internal state. Our base architecture is comprised of a ResNet18~\cite{he2016resnet} visual encoder and a 2-layer LSTM as in \cite{ddppo}.  While this model architecture is capable of nearly solving the task with \gpscompass in idealized conditions, it performs poorly without \gpscompass.  We hypothesize that learning to navigate without \gpscompass conflates two things -- learning to self-localize and learning to navigate.  In \pointnav, tracking is necessary to make progress in learning to navigate.  Insidiously, learning to navigate harms tracking as it necessarily changes the distribution of data seen by the agent. To break this cycle, we introduce an information bottleneck inspired by the work Goyal \etal~\cite{goyal2019infobot}.  Further, we augment the base architecture to improve tracking using auxiliary losses.  Our full architecture is in~[\reffig{fig:model}] and is comprised of a visual encoder, a state encoder, the information bottleneck and finally the auxiliary losses. We describe each of these components in detail below.

\csubsection{Visual-Encoder \& State-Encoder}

We use only \depth frames as this is known the perform better than using full \rgbd. As such, our Resnet-18~\cite{he2016resnet} encoder takes in two consecutive, single channel 360 x 640 depth observations and produces two 85-channel visual encodings, $v_{t-1}$ and $v_t$ for depth frames $t-1$ and $t$ receptively. $v_t$ is flattened and passed through a linear layer to produce the 384-dimensional \emph{visual embedding}, $E_v$.  Next, $v_t - v_{t-1}$ is flattened and passed through a 2-layer MLP to produce a 128-dimensional embedding vector $E_f$.

The \emph{flow embedding}, $E_f$, is the input to our ego-motion estimation calculations [\refsec{sec:background}]. The 512-dimensional concatenation $[E_v, E_f]$ is an input to our state encoder. We choose a Resnet-18~\cite{he2016resnet} due to the added simplicity of the model and negligible improvements in performance in \pointnav seen with larger visual encoders like Resnet-50~\cite{shacklett2021large}. 
One challenge in visual navigation is learning a suitable visual representation.  In the case of realistic \pointnav, this challenge is compounded by noisy observations and the more challenging environment dynamics.  As such, we first pre-train the visual encoder by training an agent for \pointnav \emph{with} \gpscompass (as the agent has \gpscompass, we do not use the information bottleneck or auxiliary losses).  After training, we take the visual encoder and freeze it to be used in our policy network.

Our state encoder is comprised by a 2-Layer LSTM. Given the 512-dimensional concatenation $[E_v, E_f]$, an embedding of the point goal, and an embedding of the previous action, we take the sum of these equal dimension vectors as is the input to our state encoder. Given this input, the LSTM state encoder outputs a belief state encoding $b_t$, as well as a new hidden state $h_t$. Our belief state encodes the agent's predicted pose relative to the PointGoal, which is then explicitly calculated in our self-tracking auxiliary losses [\refsec{sec:background}]. 
 
 \csubsection{Information Bottleneck}   
 
One of the main contributions of this work is the use of an information bottleneck to grant the agent constrained access to the ground truth \gpscompass sensor at training time. The concept of the information bottleneck was first proposed by Tishby \etal~\cite{tishby99information}, which lays out the theoretical groundings for our work. Goyal \etal~\cite{goyal2019infobot} first applied the concept of an information bottleneck to RL, where they propose that in the absence of useful reward signals, encoding limited privileged information from an oracle in a latent variable available to the agent can help guide an agent's exploration by demystifying difficult \emph{decision states}. We modify this mechanism to instead allow the training time sensor-suite to differ from the test-time sensor suite.  We construct an information bottleneck as follows:

Given the output of the LSTM at time $t$, referred to as the belief state $b_t$,
 we encode $b_t$ and point-goal updated with ground truth \gpscompass, $G$, into a latent variable $Z$, where $G$ is a 3-dimensional tensor representing the agent's pose relative to the PointGoal. We first embed the point-goal into a vector $g_t$ using a linear layer and concatenate this vector with $b_t$. The concatenation $[b_t, g_t]$ is then used as input to an MLP to predict the mean, $\mu$, and standard deviation, $\sigma$, of a normal distribution. In order to match the notation in Goyal \etal, let $S$ denote the vector $b_t$, our current belief state. We hence sample the latent variable $Z$ from the resultant distribution, $Z \sim p(Z | S, G) := N(\mu, \sigma)$\footnote{We use a normal distribution as in Goyal \etal, however any distribution with re-parameterized/differentiable sampling and a closed form expression for KL divergence would suffice.}. To encourage the agent to not rely on \gpscompass (as it won't be available at training time) we add a penalty to the objective based on the amount of information $Z$ contains. 
 \begin{align}
          \mathcal{J}(\theta)  &= 
          \mathcal{J}^\text{PPO}(\theta) - \beta \mathbb{E} \left[ D_\text{KL}(p(Z \mid G, S) \mid\mid N(0, I)) \right]
 \end{align}
where $p(Z \mid G, S)$ is the probability of a given $Z$, $D_\text{KL}$ is the Kullback–Leibler divergence, and $\mathcal{J}^\text{PPO}(\theta)$ is the objective function defined in [\refsec{sec:background}]. To illustrate the effects of the penalty term on the above objective function, let us hypothetically allow $\beta$ to approach infinity. It follows that $D_\text{KL}$ has to go to zero to avoid a large objective penalty. To achieve this, the posterior on our latent variable $Z$ has to approach the prior normal. 
Thus, $Z$ cannot encode any information about $G$ or $S$, and as such, it follows that ground-truth localization information does not reach the agent's policy (which matches evaluation conditions). 
We refer the reader to Goyal \etal for the full derivation of this training objective.  At evaluation time, $Z$ is instead sampled from the prior distribution, the standard normal distribution.

Of particular importance is the trade-off parameter $\beta$.  This controls how much information $Z$ can contain about the ground-truth \gpscompass.  We find a relative high constant value to work well in most cases, and as such $\beta{=}10$ in most of our experiments.  %

\csubsection{Tracking Auxiliary Losses}

The second important contribution of our work is our agent's use of auxiliary losses to learn self-tracking under the degradation of \gpscompass sensors. 

\xhdr{Self-tracking.} The first group of auxiliary losses are tracking losses. These aid the agent in self-localization throughout the course of a given episode.  These consist of two losses that correspond to the GPS and compass components of the \gpscompass sensor. Given the belief state at time $t$, $b_t$, we train a 2-layer MLP to predict the agent's current position and heading, $\hat{p}_t$ and $\hat{h}_t$, respectively.
Given the agent position and heading at time $t$, $p_t$ and $h_t$, respectively, we define these auxiliary losses as: 
\begin{align}
   L_\text{Position} &= \sum_t \left( \hat{p}_t - p_t \right)^2 &
   L_{\text{Heading}} &= \sum_t \text{AngularDistance}\left(\hat{h}_t, h_t\right)
\end{align}

We use the resulting position and heading predictions, $\hat{p}_t$ and $\hat{h}_t$, as inputs to the model. These predictions update the PointGoal to be relative to the agent's predicted pose, which is then then concatenated with the agent's belief state $b_t$ and used to predict the next action. We find the agent does not learn a reliable stop function without this explicit incorporation of the self-tracking prediction into the policy head.

\xhdr{Ego-motion estimation.} Inspired by the work of Datta \etal~\cite{datta2020integrating}, who used an external ego-motion estimation module to perform agent tracking, we implement ego-motion estimation as an auxiliary loss.  Our architecture takes in two consecutive frames as input and produces an embedding of the current frame and the difference of embeddings between adjacent frames, $E_f$. We train a 2-layer MLP to predict the change in position $\hat{\Delta}_{p_t}$ and heading $\hat{\Delta}_{h_t}$ given $E_f$. Given the agent's change in position, $\Delta_{p_t}$, and change in heading, $\Delta_{h_t}$, between its current and previous position, we define these auxiliary losses as:
\begin{align}
    L_{\Delta_\text{Position}} &= \sum_t \left( \hat{\Delta}_{p_t} - \Delta_{p_t} \right)^2 &
    L_{\Delta_\text{Heading}} &= \sum_t \text{AngularDistance}\left(\hat{\Delta}_{h_t}, \Delta_{h_t}\right)
\end{align}%

We use only the ego-motion auxiliary losses as learning signals. 
 
\csection{Experimental Results}
\label{sec:result}

\xhdr{Experimental Setup:}  
We consider two settings.  First, we examine performance when the agent has idealized actuation and sensing but no access to \gpscompass. This semi-idealized setting will allow us to examine the impact that constrained access to ground-truth \gpscompass has on performance.  Second, we examine performance in a more realistic setting with adding actuation noise, adding sensor noise, and `sliding'~\cite{Kadian_2020} removed.  This setting is identical to the 2020 Habitat challenge.

\csubsection{Semi-Idealized Setting Results}

\begin{wraptable}{R}{0.5\textwidth}
    \setlength{\tabcolsep}{2pt}
    \centering
    \resizebox{0.95\textwidth}{!}{
    \begin{tabular}{l c c cc c c c}
     \toprule
     & SPL & SoftSPL & Success & Dist.  \\
     \midrule
     \texttt{Baseline} & 10.95 & 68.30 & 11.47 & 1.987 \\ 
     \texttt{Datta \etal}~\cite{datta2020integrating} & 50.80 & \textbf{81.3} & 53.50 & 0.959 \\
     \texttt{AUX+IB} & 65.59 & 76.39 & 74.95 & 0.756 \\
     \texttt{AUX+IB+DSR} & \textbf{68.87} & 78.62 & \textbf{75.05} & \textbf{0.725} \\
     \bottomrule
    \end{tabular}
    }
    \caption{\textbf{Results on the semi-idealized setting validation set}. 
    }
    \label{tab:baselines-2019-setting}
\end{wraptable}

We train our agent for 300 million steps of experience for baseline comparisons. This is the same number of experience steps that Datta \etal ~\cite{datta2020integrating} use for training in their large-scale training setup, however we use 8 GPUs as opposed to 128 GPUs.  
Our agent without dynamic success reward [DSR, \refsec{sec:background}] achieves 65.59/74.95 SPL/Success and 68.87/75.05 SPL/Success with DSR  [\reftab{tab:baselines-2019-setting}, row 3 and 4], where DSR refers to the exponential term introduced in [\refsec{sec:background}]. DSR awards the agent a higher reward for calling the \texttt{STOP} action closer to the center of the PointGoal region. We observe that this simple addition yields improvements to the SPL/Success values attained by the agent and reduces the agent's average distance to the PointGoal at the end of episodes.
Compared to end-to-end learning without our proposed information bottleneck and auxiliary losses, this is 57.92 SPL higher.
Compared to
Datta \etal ~\cite{datta2020integrating}, an architecture that features an explicit odometry prediction model, our model attains +18.07 SPL and +21.55 Success. 
In [\reffig{fig:2019-ablation-scratch}] in the supplementary section we show that utilizing a pre-trained visual encoder increases SPL to 59.68 SPL at 120 million experience steps from 20.9 SPL when trained from scratch.

\begin{figure*}
\centering
  \includegraphics[width=0.45\linewidth]{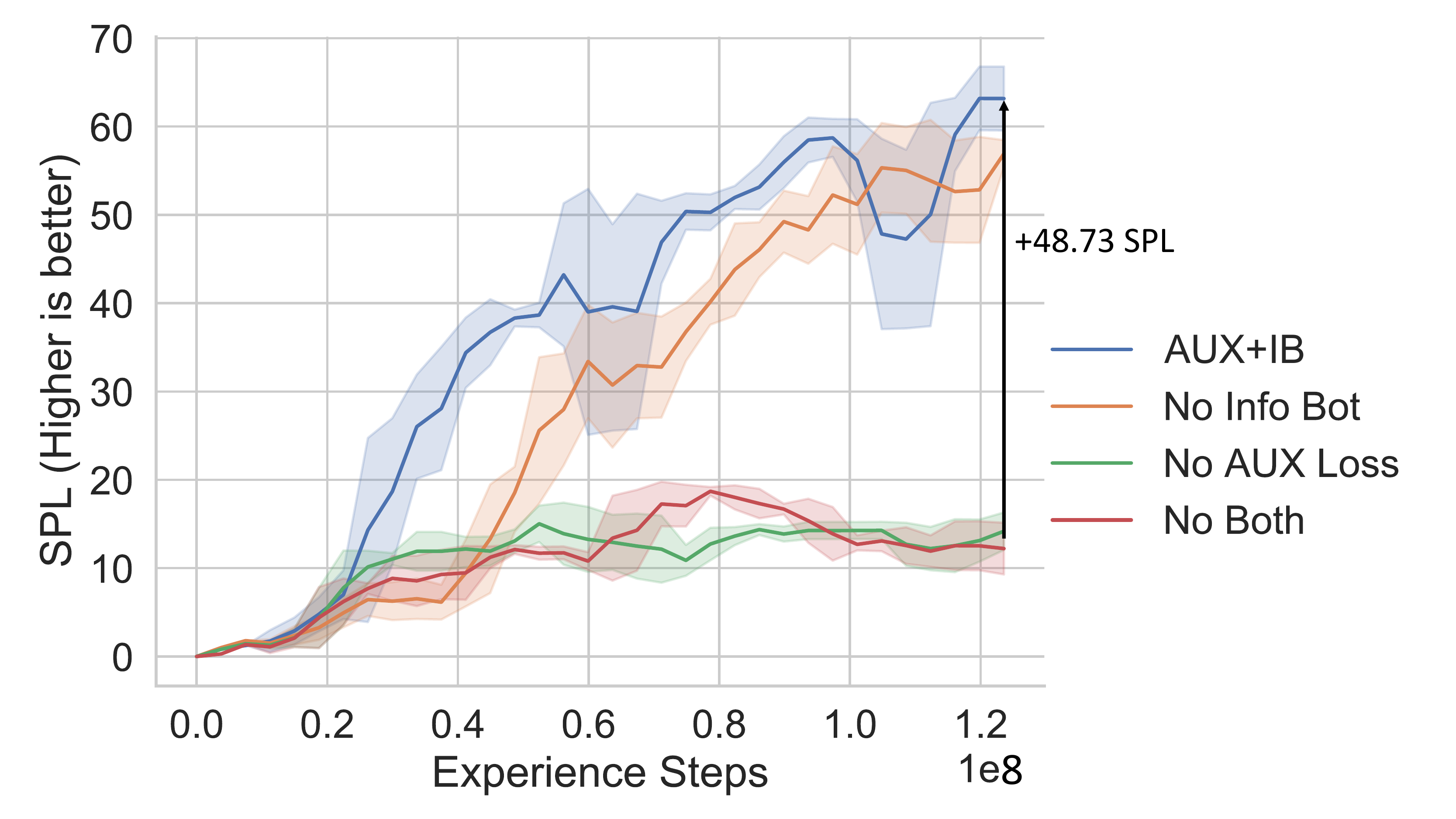}
  \includegraphics[width=0.45\linewidth]{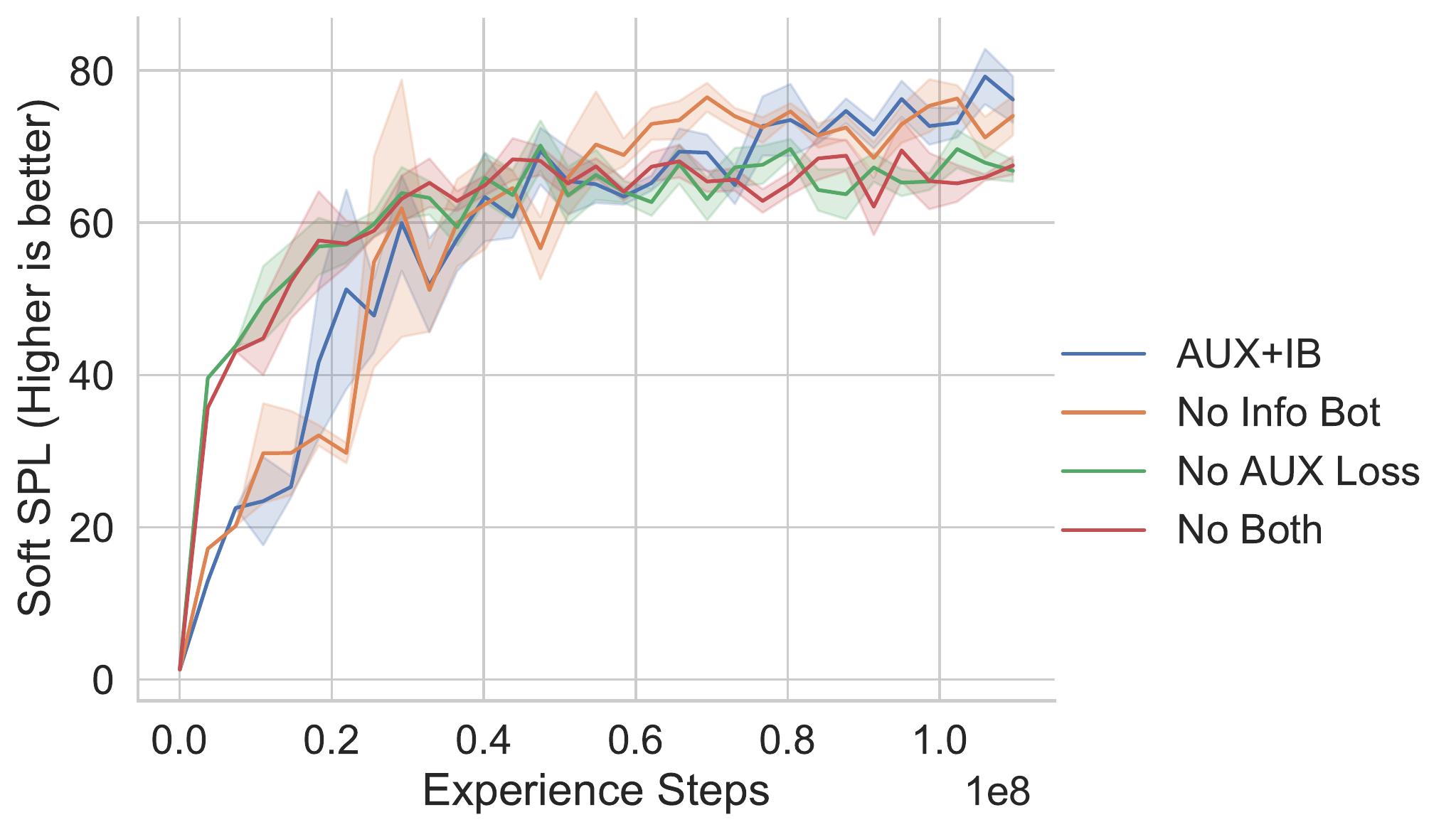}
  \caption{\textbf{Semi-Idealized Setting SPL and Soft-SPL Vs. Steps (120M Experience Steps)}
  }\label{fig:idealized-ablation}
\end{figure*}

\xhdr{Ablation Study in the Semi-Idealized Setting.} 
We compare the following, our complete architecture (AUX+IB), No Information Bottleneck  (No Info Bot), No Auxiliary Losses (No AUX Loss), and No Both, to examine the impact of our proposed contributions. To better visualize the differences between ablated agents early in training, ablations are performed to 120M experience steps without DSR [\reffig{fig:idealized-ablation}]. [\reffig{fig:idealized-ablation-full}] shows the ablation with 300M experience steps and DSR.

We find a significant performance difference between agents with access to the auxiliary losses and those without access to them, with the absolute difference between the SPL values of No Info Bot and No Aux surpassing 45 SPL [\reftab{tab:idealized-ablation}]. The same discrepancy can be observed when comparing AUX+IB to No Aux Loss and No Both. 
These results show that the agent's ability to precisely track itself is heavily tied to the availability of the auxiliary losses in training. Note that agents that do not use the auxiliary losses are still able to perform quite well in terms of Soft-SPL. This suggests that not having access auxiliary losses does not hamper the agent's ability to effectively navigate towards the goal.  However, it does heavily affect the precision of the agent's self-tracking predictions and thus its ability to stop precisely. This is illustrated by the selected trajectories presented in [\reffig{fig:trajectories}]. 

While the removal of the Information Bottleneck does not harm final performance of the agent as extensively as the removal of the Auxiliary losses, it does harm sample efficiency -- No Info Bot takes between 10 and 20 Million more experience steps to arrive at the same SPL and Success as AUX+IB [\reffig{fig:idealized-ablation}]. This suggests that having limited access to \gpscompass facilitates the agent's ability to learn, as the model can learn to use this information as a reference to correct its own predictions early on in training. This benefit is attributed to the improvement in tracking predictions resulting from the information bottleneck as seen in [\reffig{fig:aux_losses}] and [\reffig{fig:aux_losses_zoom}].

\begin{wraptable}{R}{0.5\textwidth}
    \setlength{\tabcolsep}{2pt}
    \centering
    \resizebox{0.95\textwidth}{!}{
    \begin{tabular}{l c c cc c c c}
     \toprule
     & SPL & SoftSPL & Success & Dist.  \\
     \midrule
     \texttt{Baseline} & 2.83 & 50.71 & 3.92 & 2.666 \\
     \texttt{Datta \etal}~\cite{datta2020integrating} & 4.70 & 57.60 & 6.00 & 1.843 \\
     \texttt{AUX+IB} & 16.59 & 55.36 & 23.14 & 1.575 \\
     \texttt{AUX+IB+DSR} & \textbf{19.86} & \textbf{60.20} & \textbf{26.66} & \textbf{1.474} \\
     \bottomrule
    \end{tabular}}
    \caption{\textbf{Baselines on the realistic setting (Habitat Challenge 2020) validation set}. %
    }
    \label{tab:baselines-2020-setting}
\end{wraptable}

\csubsection{Realistic Setting Results}

We expect the transition to realistic sensing and actuation to result in a challenging learning problem. Thus, we extend our training time to 350 million steps of experience for all experiments.

Under these harsh conditions, our model with DSR achieves 16.59 SPL, 23.14 Success, and 55.36 SoftSPL and  19.86 SPL, 26.66 Success, and 60.20 Soft SPL with DSR [\reftab{tab:baselines-2020-setting}, row 3 and 4].  For comparison, the baseline achieves 2.83 SPL and 3.92 Success.  We again see that Soft-SPL is considerably more forgiving to less precise tracking and the baseline achieves 50.71 Soft-SPL.  Compared to Datta \etal~\cite{datta2020integrating}\footnote{Performance metrics reported in Datta \etal~\cite{datta2020integrating} were obtained under noisy actuations but perfect sensing.}, we achieve +15.41 SPL and +20.66 Success.  Unlike in the semi-idealized setting, we see that our model narrowly outperforms Datta \etal~\cite{datta2020integrating} in Soft-SPL, achieving +2.60 [\reftab{tab:baselines-2020-setting}].

\xhdr{Ablation Study in Realistic Setting.} We perform an ablation study analogous to the one performed for the semi-idealized setting on the realistic (Habitat Challenge 2020) setting to gauge the effects of each of our contributions. Again we observe a large performance gap between models with access to auxiliary losses, and those without them [\reffig{fig: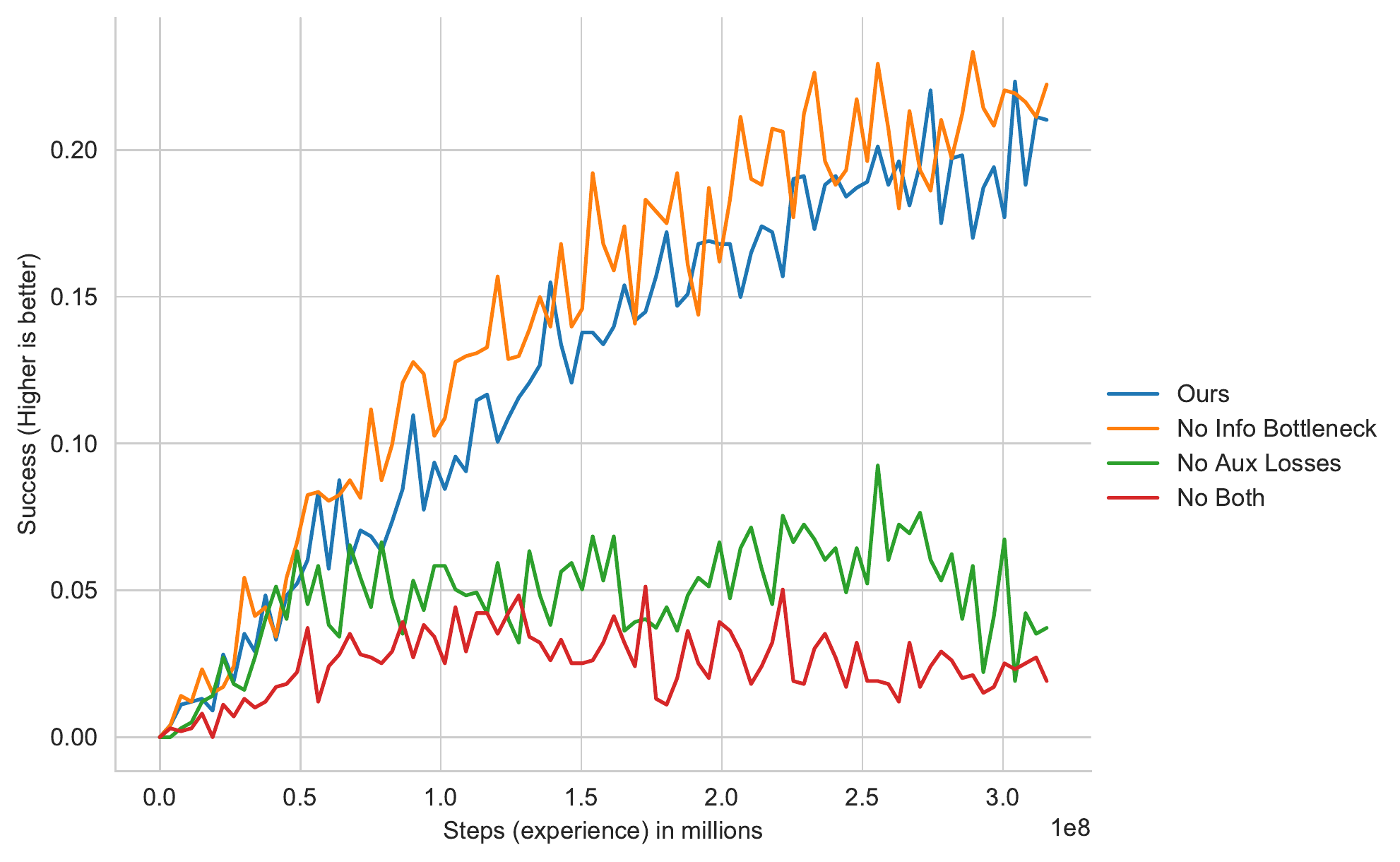}]. AUX+IB and No Info Bot reach SPL values of 19.86 and 16.55 respectively. This again shows that the auxiliary losses presented are key to enabling effective and precise agent tracking, which is fundamental to episodic success. All ablation agents perform comparably well in terms of Soft-SPL, indicating that precise self-tracking is not required to navigate towards the goal but is to stop in the correct location.

\begin{wraptable}{R}{0.5\textwidth}
    \setlength{\tabcolsep}{2pt}
    \centering
\resizebox{0.95\textwidth}{!}{    \begin{tabular}{l c c cc c c c}
     \toprule
     & SPL & SoftSPL & Success & Dist. & Exec. Time  \\
     \midrule
     \texttt{RGBD+DDPPO Baseline} & 0.10 & 3.80 & 0.30 & 6.93 & -\\
     \texttt{Datta \etal}~\cite{datta2020integrating} & 11.9 & 58.6 & 15.7 & 2.23 & 2h 18m \\
     \texttt{AUX+IB+DSR} & 12.2 & 56.1 & 16.3 & 2.08 & 2h 42m \\
     \midrule
      \texttt{Occupancy Anticipation}~\cite{ramakrishnan2020occupancy} & 22.0 & 47.0 & 29.0 & 2.57 & 11h 4m \\
      \texttt{DAN}~\cite{DAN} & 37.7 & 52.1 & 64.5 & 0.69 & 1d 13h 29m \\
      \texttt{VO} & 52.5 & 66.5 & 71.7 & 0.80 & 5h 49m \\
     \bottomrule
    \end{tabular}}
    \caption{\textbf{Results on the realistic setting (Habitat Challenge 2020) test set}. 
    }
    \label{tab:habitat-challenge-2020-results}
\end{wraptable}

\xhdr{Online leaderboard results.}  For the online Habitat Challenge 2020 leaderboard test set [\reftab{tab:habitat-challenge-2020-results}], our model performs considerably better than the DD-PPO~\cite{ddppo} baseline model, achieving 12.2 SPL and 16.3 Success vs. 0.1 SPL and 0.3 Success.  Here we see a trend similar to performance on the semi-idealized setting between our model and Datta \etal. While our model outperforms Datta \etal in terms of SPL and Success (+0.3 and +0.6 respectively), Datta \etal fares better in terms of Soft-SPL (-2.5). Our model also shows the best performance with respect to final distance to goal when compared to Datta \etal and Occupancy Anticipation \cite{ramakrishnan2020occupancy}. We note that the state of the art in realistic PointGoal navigation has advanced rapidly in parallel to this work, with an approach reaching SPL, Success and Soft-SPL values of 52.5, 66.5 and 71.7 respectively currently topping the leaderboard. A second approach, Differentiable SLAM-net~\cite{DAN} achieved SPL, Success and Soft-SPL values of 37.7, 52.1 and 64.5 respectively. While both these works attain superior performance to our approach, it is important to note they are much more computationally expensive at inference time methods. This can be observed in the execution time column of [\reftab{tab:habitat-challenge-2020-results}].  Inference time is an important consideration for deployment as it limits how quickly a robot can perform the task and motivates further research in the direction of end-to-end policy networks for realistic PointGoal Navigation.

\csection{Analysis of Information Bottleneck}
\label{sec:ib_analysis}

\begin{figure*}
\centering
  \includegraphics[width=0.45\linewidth]{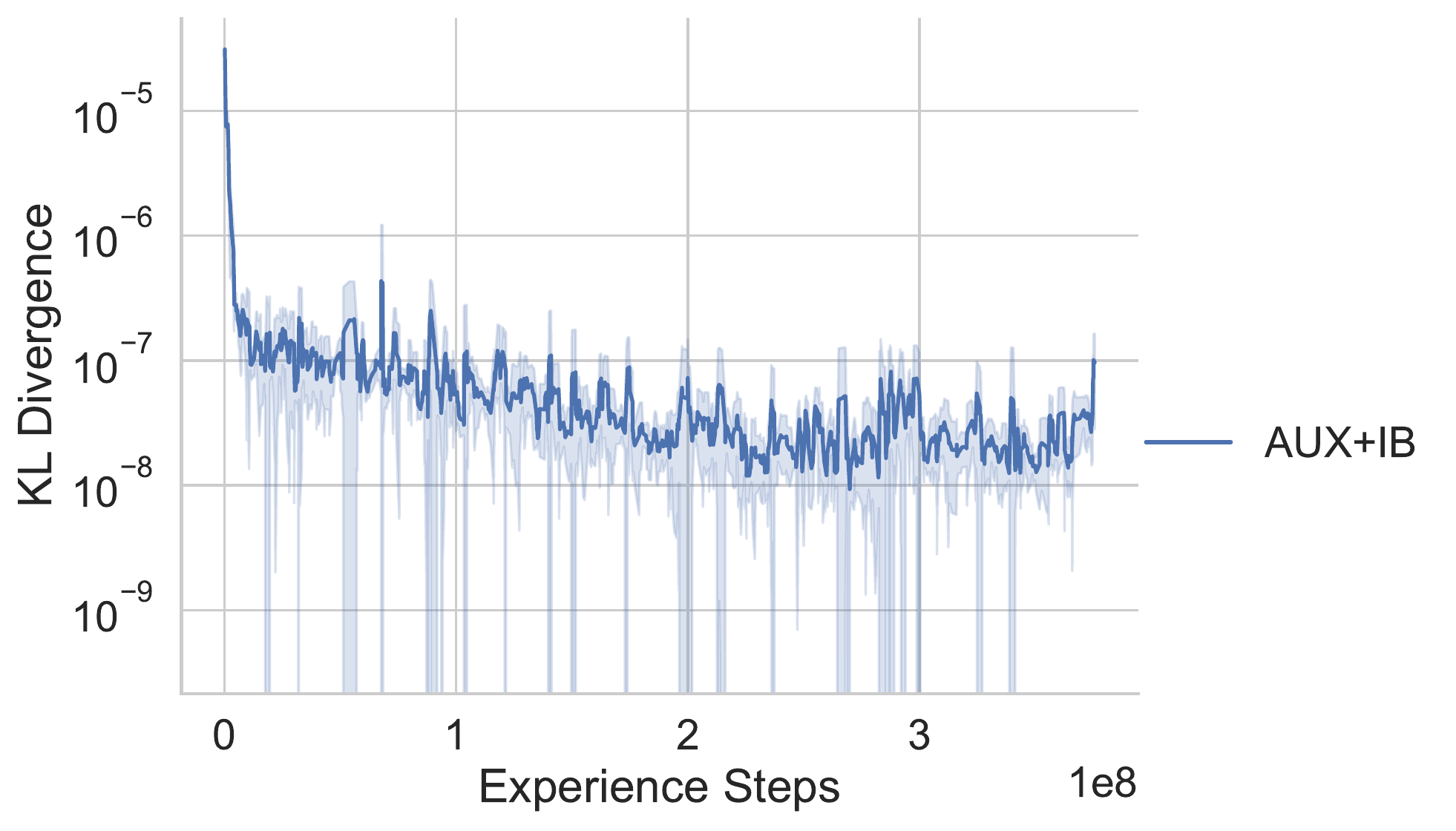}
  \includegraphics[width=0.45\linewidth]{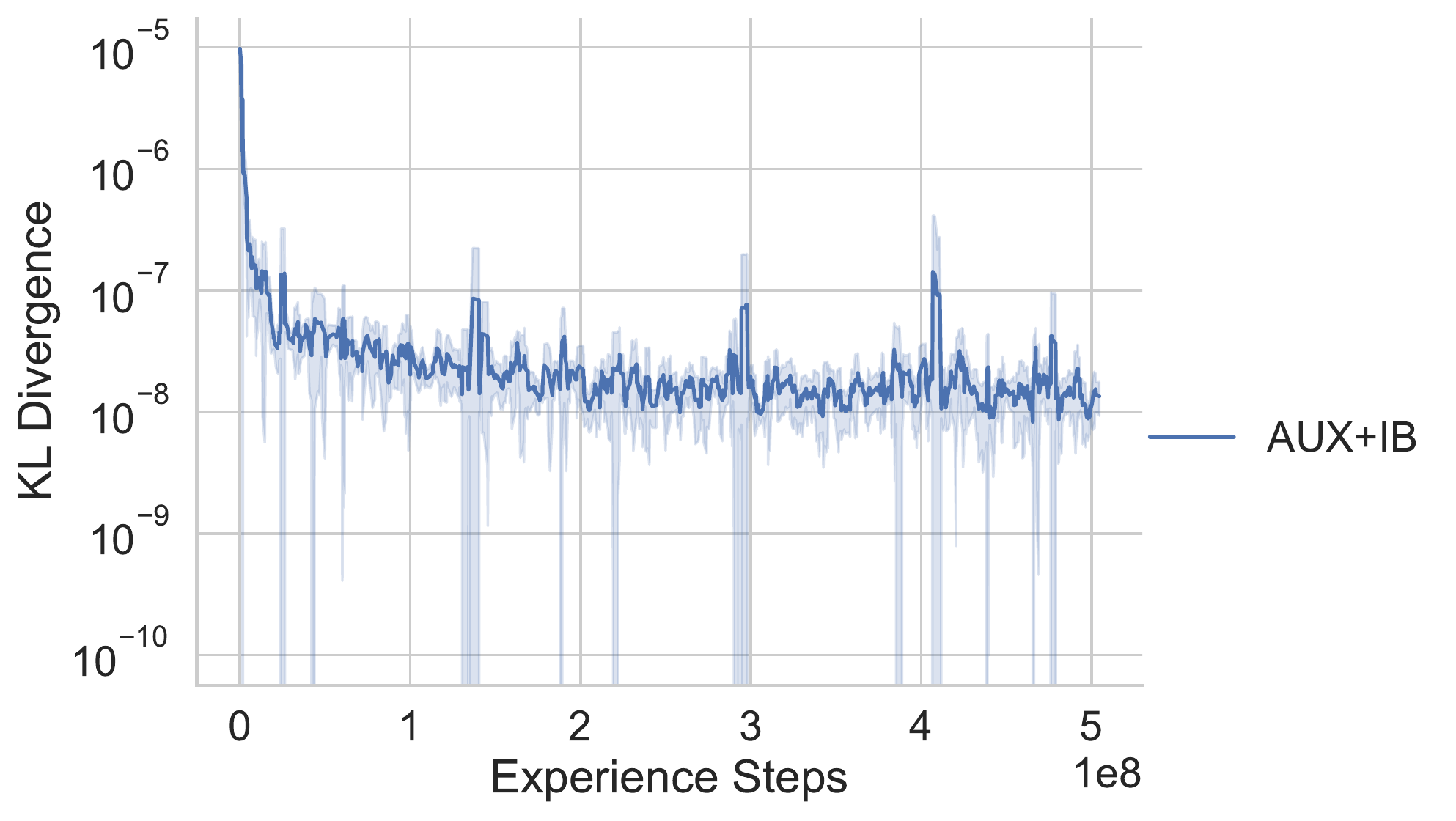}
  \caption{Semi-Idealized (left) and Realistic (right) Setting KL Divergence between $p(Z | S, G)$ and $N(0, I)$ prior Vs. Steps ($\beta = 10.0$, DSR).}\label{fig:kl_div}
\end{figure*}

\xhdr{Effects of Beta on KL-Divergence:} Despite using a fixed beta value to penalize the use of privileged information throughout training, we observe that the agent learns to reduce it's reliance on the privileged information as training progresses. This is illustrated by the decreasing KL divergence observed between the latent distribution $p(Z | S, G)$ and the prior $N(0, I)$ under a fixed $\beta = 10$ depicted in \reffig{fig:kl_div}. We find that using too small of a fixed beta value enables the agent overuse the privileged information at it's discretion, resulting in failure to generalize at evaluation time [\reffig{fig:beta_1}].

Similarly, beta annealing strategies resulted in failure to generalize. We found large discrepancies between the agent’s behavior during beta decay and after the final beta value is achieved, indicating that the policy learned with low beta is unable to adapt to a high beta regime [\reffig{fig:beta_5e-3}]. While a large, fixed beta value prevents this lack of generalization from taking place, we believe that better performance can be achieved through improved annealing strategies in future research.  

\csection{Conclusion}
\label{sec:conclusion}

In this work we propose a novel reinforcement learning architecture capable of performing well in embodied \pointnavfull both in semi-idealized and realistic simulated scenes. Our architecture leverages an information bottleneck to grant limited access to ground truth Compass+GPS, allowing the agent to benefit from localization at training time. We find that the information bottleneck improves sample efficiently, allowing the agent to reach competitive performance in terms of Success and SPL with fewer experience steps, while also improving performance.  We additionally propose a set of auxiliary losses to promote self-tracking in our agent and show through experimentation that these auxiliary losses are key to succeed in \pointnavfull. We provide strong experimental results for our architecture both in semi-idealized (Habitat Challenge 2019 \emph{without} Compass+GPS) and realistic (Habitat Challenge 2020) settings. Specifically, our method out performs existing baselines on the semi-idealized setting by +18.07 SPL and +15.16 SPL on the realistic setting.

\bibliography{bib/strings.bib,bib/main.bib}

\newpage
 \appendix
\renewcommand\thesection{\Alph{section}}
\setcounter{section}{0}
\renewcommand\thefigure{A\arabic{figure}}
\renewcommand\thetable{A\arabic{table}}
\setcounter{figure}{0}
\setcounter{table}{0}
\phantomsection

  \section{Evaluation Metrics}
\label{sec:metrics}
In this section we provide the formal definitions for the PointGoal Navigation success metrics presented in Section 3. As previously stated, performance is evaluated using four key metrics: Success, SPL~\cite{anderson2018evaluation}, Soft-SPL~\cite{datta2020integrating}, and Distance to Goal. Distance to Goal measures the agent's geodesic distance to the target when the \texttt{stop}  action was called. Success indicates whether the agent predicted \texttt{stop} within $R_\text{PG}$ of the goal, where $R_\text{PG} = 0.36$ meters in all of our experiments. SPL weighs the Success value by the length of the agent's path compared to the optimal path. Soft-SPL replaces the binary Success indicator with percent progress towards goal.

Formally, let $s_T$ be the coordinate of the agent when \texttt{stop} was called, $g$ be the goal coordinate, and $d_\text{init}$ be the initial geodesic distance to goal, and $p$ be the length of the path taken by the agent, then
\begin{align}
    \text{DistToGoal} &= \text{GeoDist}(s_T, g) \\
    \text{Success} &= \mathbf{1}[[{\text{DistToGoal} \leq R_{PG}}]] \\
    \text{SPL} &= \text{Success} \cdot \left( \frac{d_\text{init}}{\max{(d_\text{init}, p)}} \right) \\
    \text{Soft-SPL} &= \left(1 - \frac{\text{DistToGoal}}{d_\text{init}} \right) \cdot \left( \frac{d_\text{init}}{\max{(d_\text{init}, p)}} \right)
\end{align}
Note that throughout this paper, we will report Success, SPL and Soft-SPL as percentages for exposition clarity. 

Throughout this work we focus on SPL and soft-SPL given our focus on serializing learning self-localization and navigation. Intuitively, SPL can be understood as a measure of an agents self-localization ability, while Soft-SPL serves as a measure of the quality of a navigation policy. This is because in order to achieve a non-zero SPL, an agent is required to accurately predict its position relative to the goal coordinate and call the \texttt{stop} action accordingly. As such, achieving a high SPL score is more heavily dependent on an agent's ability to accurately localize itself throughout an episode. Soft-SPL on the other hand does not have this stringent self-localization requirement. Instead, Soft-SPL gauges how much closer the agent is to the goal coordinate at the end of an episode relative to its distance from the goal at the beginning of the same episode. As such, Soft-SPL informs researchers of the quality of a navigation policy even for unsuccessful episodes, providing a more holistic understanding of an agent's navigation ability compared to SPL. 
 
 \section{Implementation Details}
\label{sec:supplementary}
 
In this section we provide implementation details on our proposed architecture as a supplement to Section 4.

\subsection{Visual-Encoder}

As mentioned in Section 4, our ResNet-18~\cite{he2016resnet} encoder takes in two consecutive, single channel 360 by 640 depth observations and produces two fixed-length embeddings. 
Each depth observation is passed through an average pool with kernel size 3 and passed through a ResNet-18 backbone to produce a visual encoding. Note, the ResNet-18 visual encoder is pretrained as mentioned in Section 5.  The output of the pre-trained and frozen ResNet-18  backbone is then passed through two additional basic residual blocks to reduce the channel dimension to 85.
Let $v_t$ denote the visual encoding for the current frame and $v_{t-1}$ denote the previous frame visual encoding.

The visual encodings $v_{t}$ and $v_{t-1}$ are then used to generate the embeddings. First, $v_t$ is flattened and pass thought a 1-layer MLP with LayerNorm and ReLU to produce a 384 dim vector $E_v$.  Next, $v_t - v_{t-1}$ is flattened and passed through a 2-layer MLP with LayerNorm and ReLU to produce a 128 dim vector $E_f$ -- $E_f$ is set to the zero vector when $t=0$.  $E_f$ is used as the input to the ego-motion auxiliary loss calculations mentioned in Section 5.

\subsection{State-Encoder}
Our state encoder is comprised by a 2-Layer LSTM with \texttt{input size} = 512 and \texttt{hidden dim} = 512. Here, the \texttt{visual embedding} $E_v$ and the \texttt{flow embedding} $E_f$, as well as a \texttt{context embedding}, $E_c$ composed of an embedding of the PointGoal and an embedding of the previous action are used as inputs. The vectors $E_v$ and $E_f$ are computed as specified in Section 8.1, and and are concatenated together, where $[E_v, E_f]$ is a 512-d vector. Then the PointGoal observation (goal coordinate relative to the agent's \emph{initial position}) is passed through a linear layer to produce a 512-d vector representation of the PointGoal. Next, the previous action performed by the agent is encoded into a 512-d vector using a word embedding layer. The PointGoal embedding and the previous action embedding are then summed with $[E_v, E_f]$ to produce the LSTM input. 

The LSTM state encoder outputs a 512-d belief state $b_t$. The belief state is used as the input in our self-tracking auxiliary losses (Section 5). 
 
\subsection{Information Bottleneck}   
 
Given the predicted belief state $b_t$, we encode $b_t$ and the PointGoal updated with ground truth \gpscompass, $G$, into a latent variable $Z$. We first embed the ground truth PointGoal observation into a 32-d vector $g_t$ using a lineear layer and concatenate this vector with $b_t$.  The resulting concatenated vector $[b_t, g_t]$ of dimension 544 is then used as input to a 2-layer MLP with a hidden size of 128, a LayerNorm + ReLU activation (note that the final linear layer does not use an activation), and an output size of 64.

The 64-d output is then split into $\mu$ and standard deviation $\sigma$ of a normal distribution.  We use a softplus activation on $\sigma$ and no activation on $\mu$.  $Z$ is then sampled from $N(\mu, \sigma)$.

During evaluation, $G$ is unknown and thus we set $\mu$ to the zero vector and $\sigma$ to the ones vector as we use the standard normal distribution as the prior in the information bottleneck.

Using the output from the self-tracking auxiliary loss, we update the PointGoal to be relative to the current agent location.  We then pass this through a linear layer to create a 32-d vector and concatenate with $Z$.  The 64-d vector is then compressed to 32-d with a linear layer and ReLU.

To encourage the agent to not rely on \gpscompass (as it won't be available at training time) we add an penalty to the objective based on the amount of information $Z$ contains.  Specifically, we use the same training objective as  presented in Goyal \etal
 \begin{align}
          \mathcal{J}(\theta)  &=  \mathcal{J}^\text{PPO}(\theta) - \mathbb{E} \left[ \beta I(p(Z | G, S);N(0, I)) \right] \\
          & \leq \mathcal{J}^\text{PPO}(\theta) - \mathbb{E} \left[\beta D_\text{KL}(p(Z | G, S) | N(0, I)) \right]
 \end{align}
where $G$ is the ground-truth \gpscompass, $S$ is our belief state, $I(Z;N(0, I))$ is the mutual information between latent variable $Z$ and the standard normal distribution, $D_\text{KL}$ is the Kullback–Leibler divergence and $\mathcal{J}^\text{PPO}(\theta)$ is the objective function defined in Section 3.

In practice, we find that the gradients for the KL divergence term can be highly correlated leading to worse performance.  To decorate, we random subsample to 20\% before taking the mean, we denote this as \texttt{subsampled\_mean}.

\subsection{Tracking Auxiliary Losses}

\xhdr{Ego-motion estimation.} We use the \texttt{flow embedding}, $E_f$ calculated in Section 8.1 to predict ego-motion as an auxiliary loss. To do this, we first pass $E_f$ through a single linear layer in order to convert $E_f$ into a 3 dimenasional tensor representative of the change in pose of the agent, $\Delta_{\text{ego-motion}} = [\Delta_{x}, \Delta_{y}, \Delta_{\theta}]$. Let $\hat{\Delta}_{P}$ denote the predicted ego-motion delta for position and let $\hat{\Delta}_{H}$ denote the ego-motion delta for heading.   Next, we obtain the ground truth \texttt{GPS} reading for frames $t$ and $t - 1$, as well as the ground truth \texttt{Compass} readings for frames $t$ and $t - 1$ from the environment. We compute the ground truth deltas as $\Delta_{P} = \texttt{GPS}_t - \texttt{GPS}_{t-1}$ and $\Delta_{H} = \texttt{Compass}_t - \texttt{Compass}_{t-1}$. Given the aforementioned deltas, we define these auxiliary losses as:

\begin{align}
    L_{\Delta_\text{Position}} &= \texttt{subsampled\_mean} \left[\left( \hat{\Delta}_{P} - \Delta_{P} \right)^2 \right]
\end{align}

\begin{align}
     L_{\Delta_\text{Heading}} = \texttt{subsampled\_mean} \left[\texttt{angular\_distance}\left(\hat{\Delta}_H, \Delta_H\right) \right]
\end{align}

\xhdr{Self-tracking.} We use the state encoding $b_t$ computed in Section 8.2 to calculate self-tracking losses. The first loss, Position loss, is defined as the MSE loss of the agent's predicted position and the ground truth position. This is computed by first passing the state encoding through a two-layer MLP with at 256-d hidden dimension and a ReLU hidden activation.

The \texttt{GPS\_head} thus decodes the predicted GPS reading $\hat{P} = [\hat{x}, \hat{y}]$ from the 512-d state encoding $b_t$. Next, given the ground truth GPS reading $P$, we compute the \texttt{GPS\_loss},  $L_\text{Position}$ as follows:

\begin{align}
   L_\text{Position} &= \texttt{subsampled\_mean} \left[\left( \hat{P} - P \right)^2 \right]
\end{align}

The second tracking loss, Heading Loss, is defined as the angular distance between the agent's predicted compass heading and the ground truth heading across. This is computed by first passing the state encoding through a second two layer MLP.

The \texttt{Compass\_head} thus decodes the predicted Compass reading $(\hat{H} = \hat{\theta})$ from the 512-d state encoding $b_t$. Next, given the ground truth Compass reading $H$, we compute the \texttt{Compass\_loss},  $L_\text{Heading}$ as follows:

\begin{align}
     L_{\text{Heading}} = \texttt{subsampled\_mean} \left[\texttt{angular\_distance}\left(\hat{H}, H\right) \right]
\end{align}

\begin{table}
    \setlength{\tabcolsep}{2pt}
    \centering
    \begin{tabular}{l c c cc c c c}
     \toprule
     & Aux.~Losses & IB && SoftSPL & SPL & Success  \\
     \midrule
     \texttt{1} & - & - && 68.30 & 10.95 & 11.47 \\
     \texttt{2} & \checkmark & - && \textbf{77.24} & 58.29 & 63.08 \\
     \texttt{3} & - & \checkmark &&
     69.95 & 14.17 & 15.19 \\
     \texttt{4} & \checkmark & \checkmark && 74.06 & \textbf{59.68} & \textbf{67.20} \\
     \bottomrule
    \end{tabular}
    \caption{\textbf{Ablation results on the semi-idealized setting}  (120M Frames). 
    From the results of this ablation study it can be observed that much of our agent's ability to perform well in terms of Success and SPL is closely tied to the availability of the auxiliary losses. As such, we infer that self-tracking performance is attributed almost exclusively to the auxiliary losses. The information bottleneck marginally increases the performance of our agent in both SPL and Success in idealized conditions, at the expense of Soft-SPL. 
    }
    \label{tab:idealized-ablation}
\end{table}

\begin{figure*}
\centering
  \includegraphics[width=0.45\linewidth]{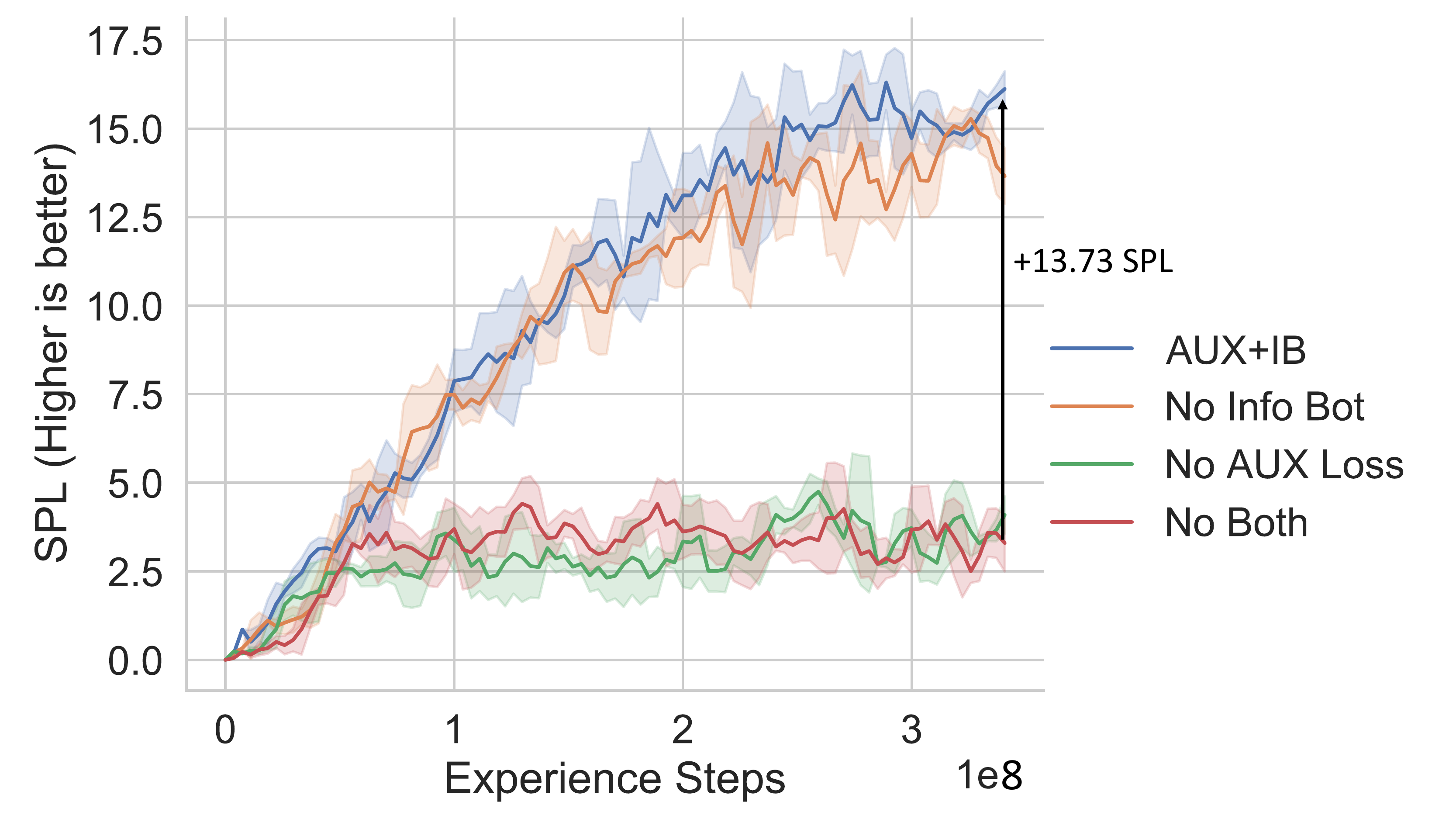}
  \includegraphics[width=0.45\linewidth]{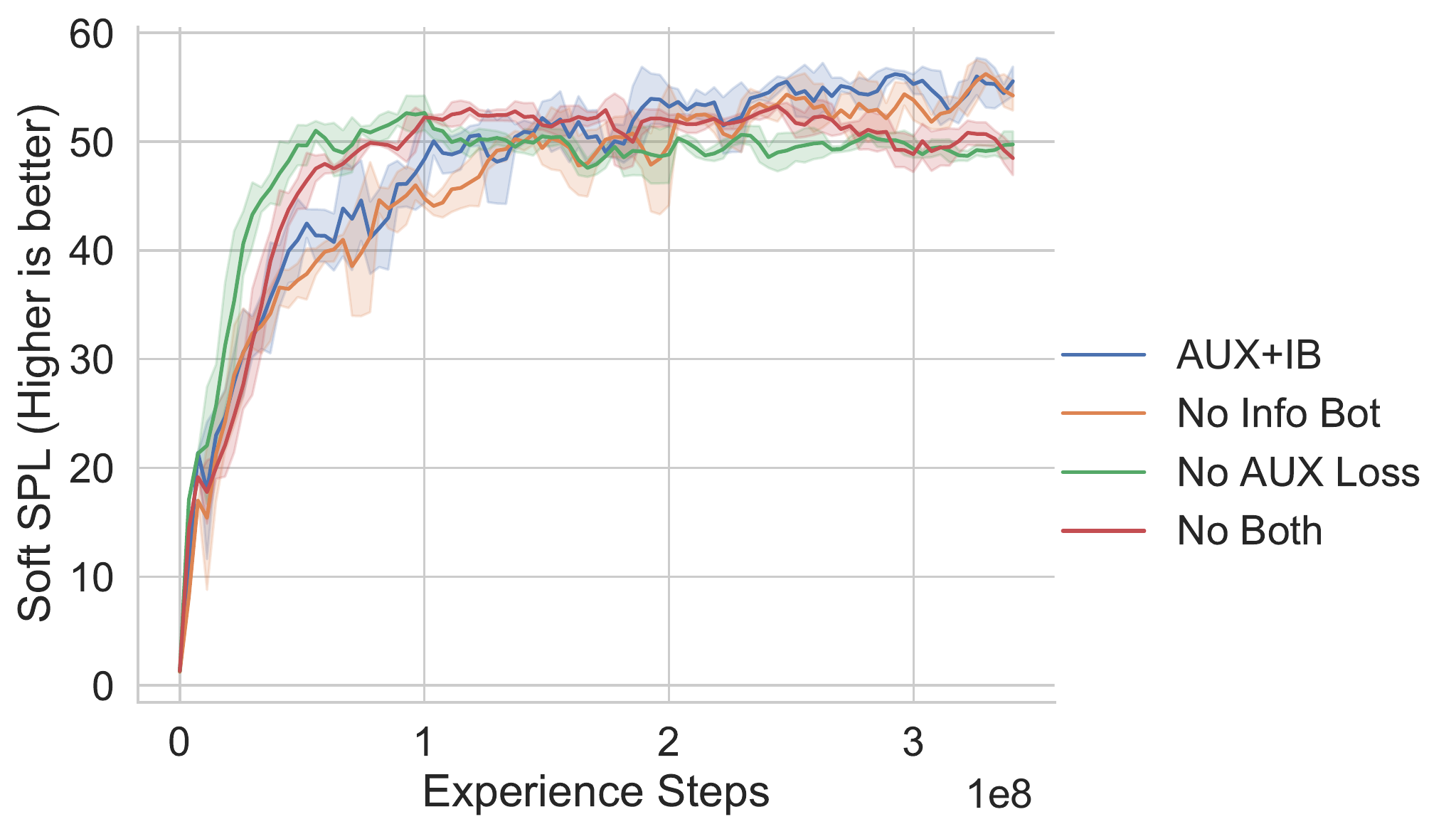}
  \caption{\textbf{Realistic setting SPL and Soft-SPL Vs. Steps (300M Experience Steps - DSR)}. 
  Our ablation studies reveal a similar trend to the ablation study on the semi-idealized setting. Agents which include our proposed auxiliary losses showcase much higher performance in terms SPL and Success, indicating improved self-tracking precision. We also note that the inclusion of the Info Bot marginally increases the performance in terms of SPL when comparing AUX+IB to No Info Bot. Soft-SPL remains comparable between all agents given that Soft-SPL is insensitive to precise self-tracking. Curves in this figure are smoothed using a windowed average over three values.
  }
  \label{fig:figures/2020_ablation_success_vs_steps.pdf}
\end{figure*}  

\begin{figure}
\centering
  \includegraphics[width=0.45\linewidth]{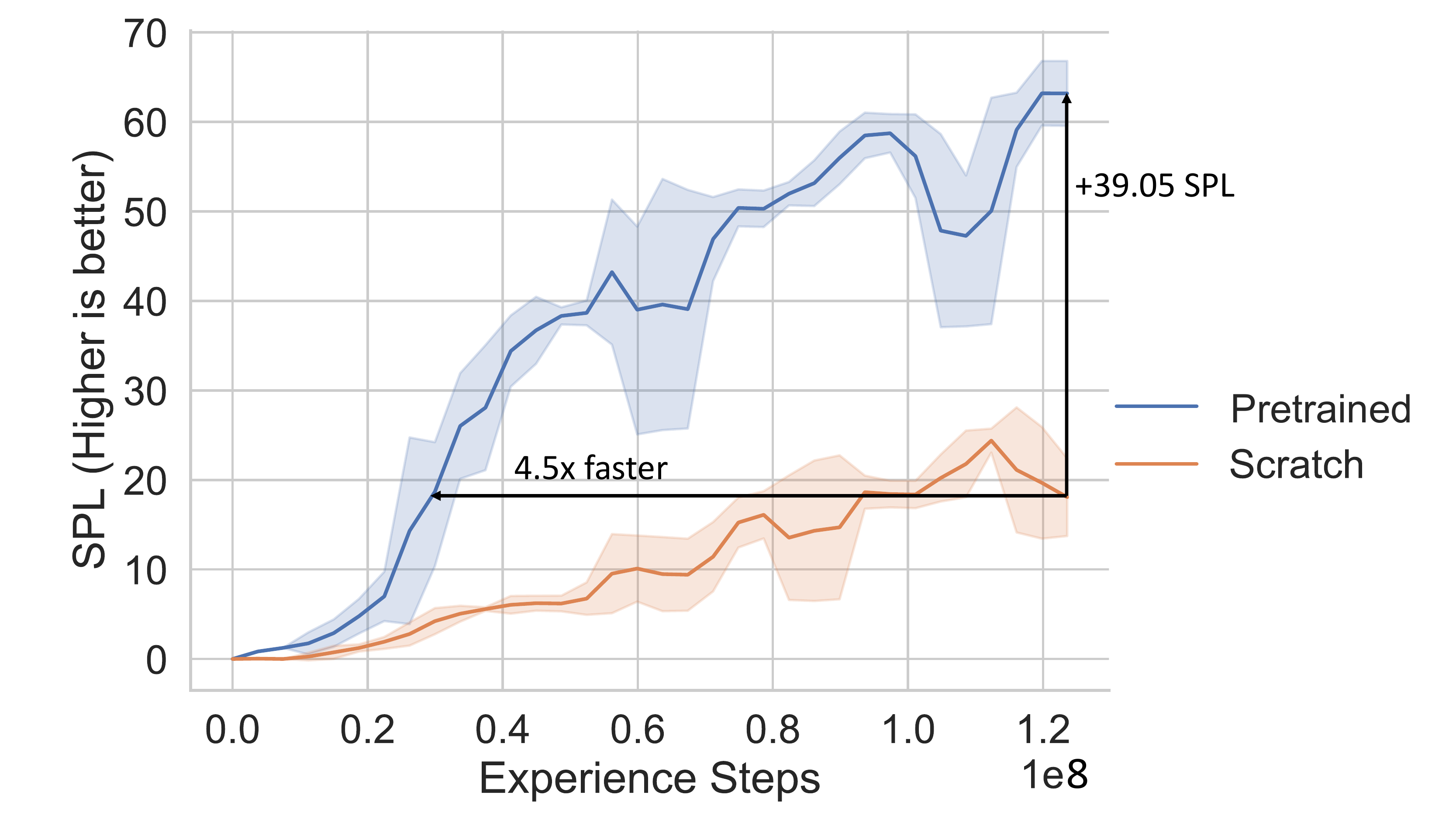}
  \includegraphics[width=0.45\linewidth]{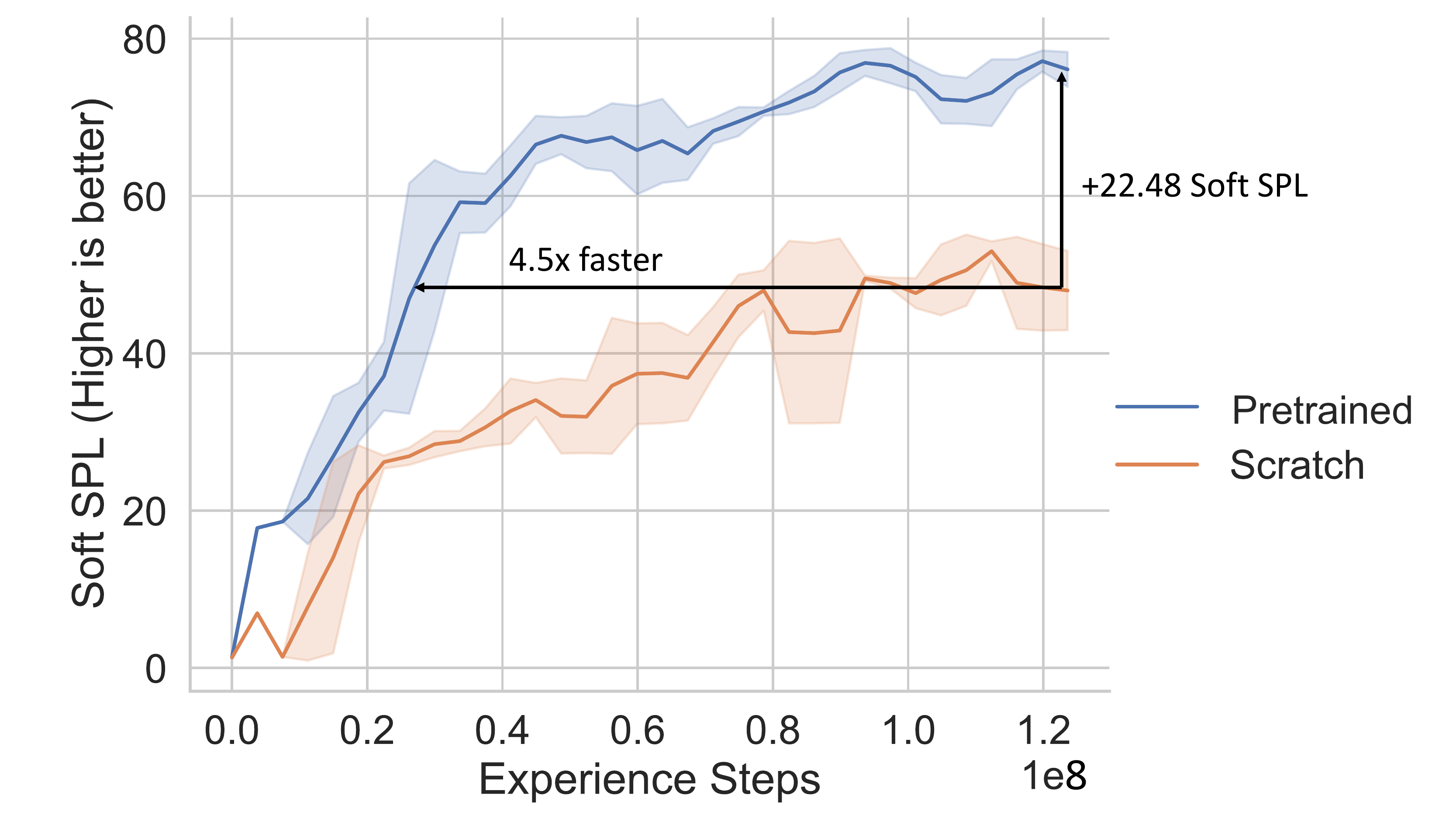}
 \caption{\textbf{Semi-Idealized SPL and Soft-SPL with pretrained encoder vs. training from scratch}. Adding a pretrained encoder to the model results in superior performance throughout training in terms of both SPL and Soft-SPL. The pretrained visual encoder provides the agent with a visual representation already amenable to learning navigation, allowing it to more quickly pickup on action sequences that lead to high reward. Curves in this figure are smoothed using a windowed average over three values.}\label{fig:2019-ablation-scratch}
\end{figure}

\begin{figure*}
\centering
  \includegraphics[width=0.45\linewidth]{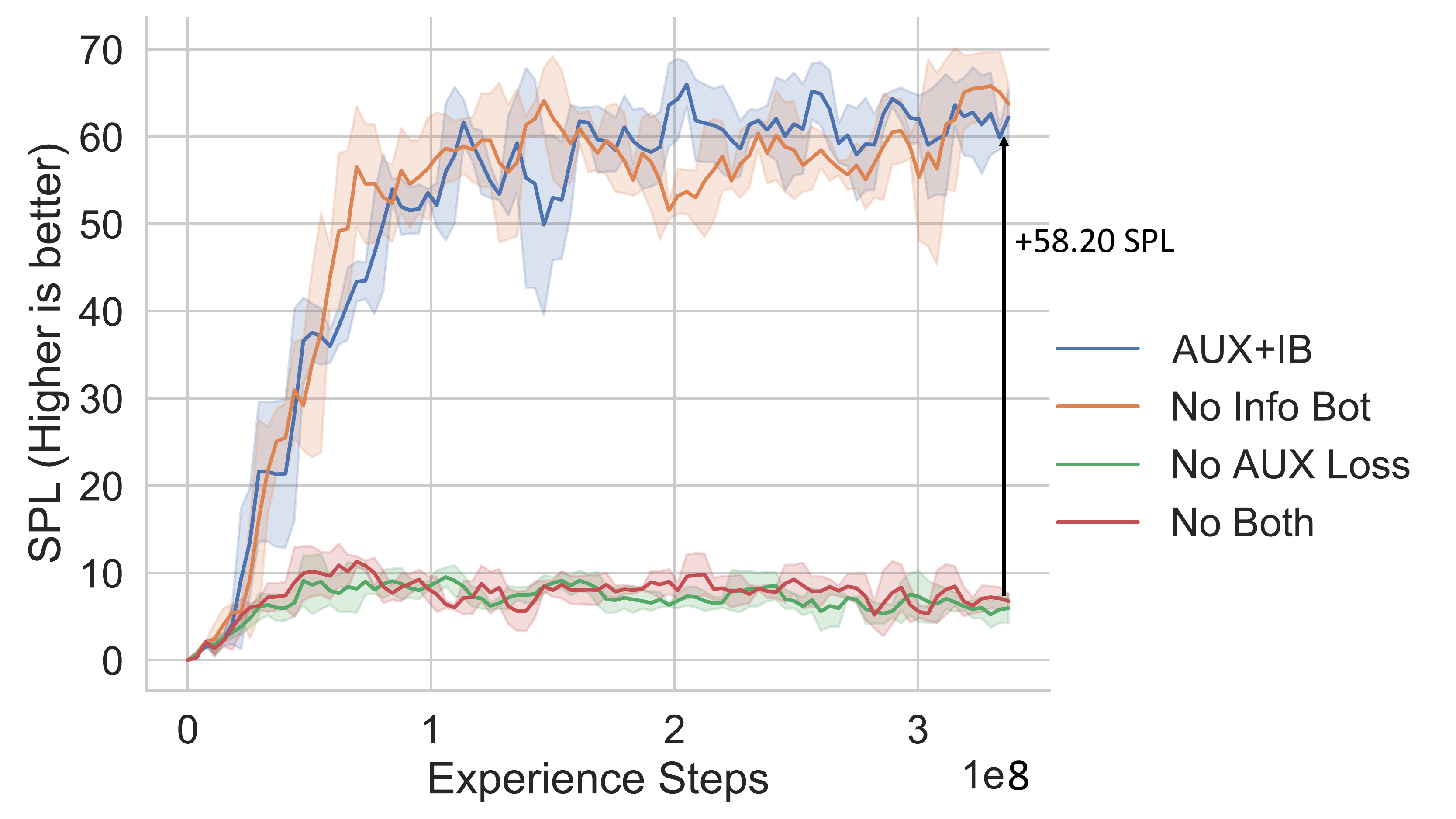}
  \includegraphics[width=0.45\linewidth]{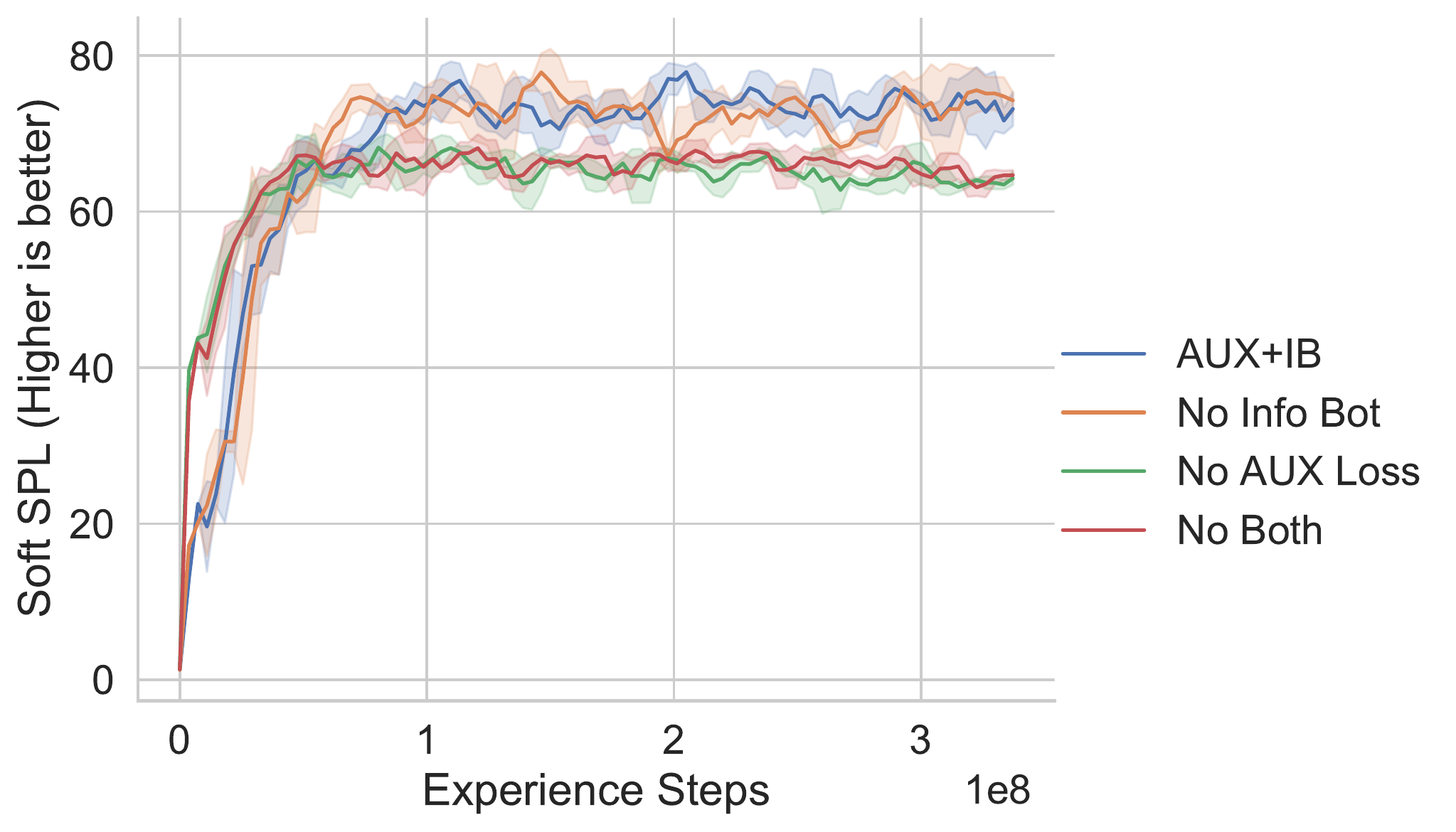}
  \caption{\textbf{Semi-Idealized Setting SPL and Soft-SPL Vs. Steps (300M Experience Steps - DSR)}. While the full ablation curves do not illustrate gains in sample efficiency from the Info Bot as well, they serve to illustrate our methods full potential in terms of SPL, achieve +58.20 SPL relative to the baseline. Additionally, we find that adding DSR in the semi-idealized setting reduces the sample efficiency gaps between No Info Bot and AUX+IB. Curves in this figure are smoothed using a windowed average over three values.}\label{fig:idealized-ablation-full}
\end{figure*}  

\begin{figure*}
\centering
  \includegraphics[width=0.45\linewidth]{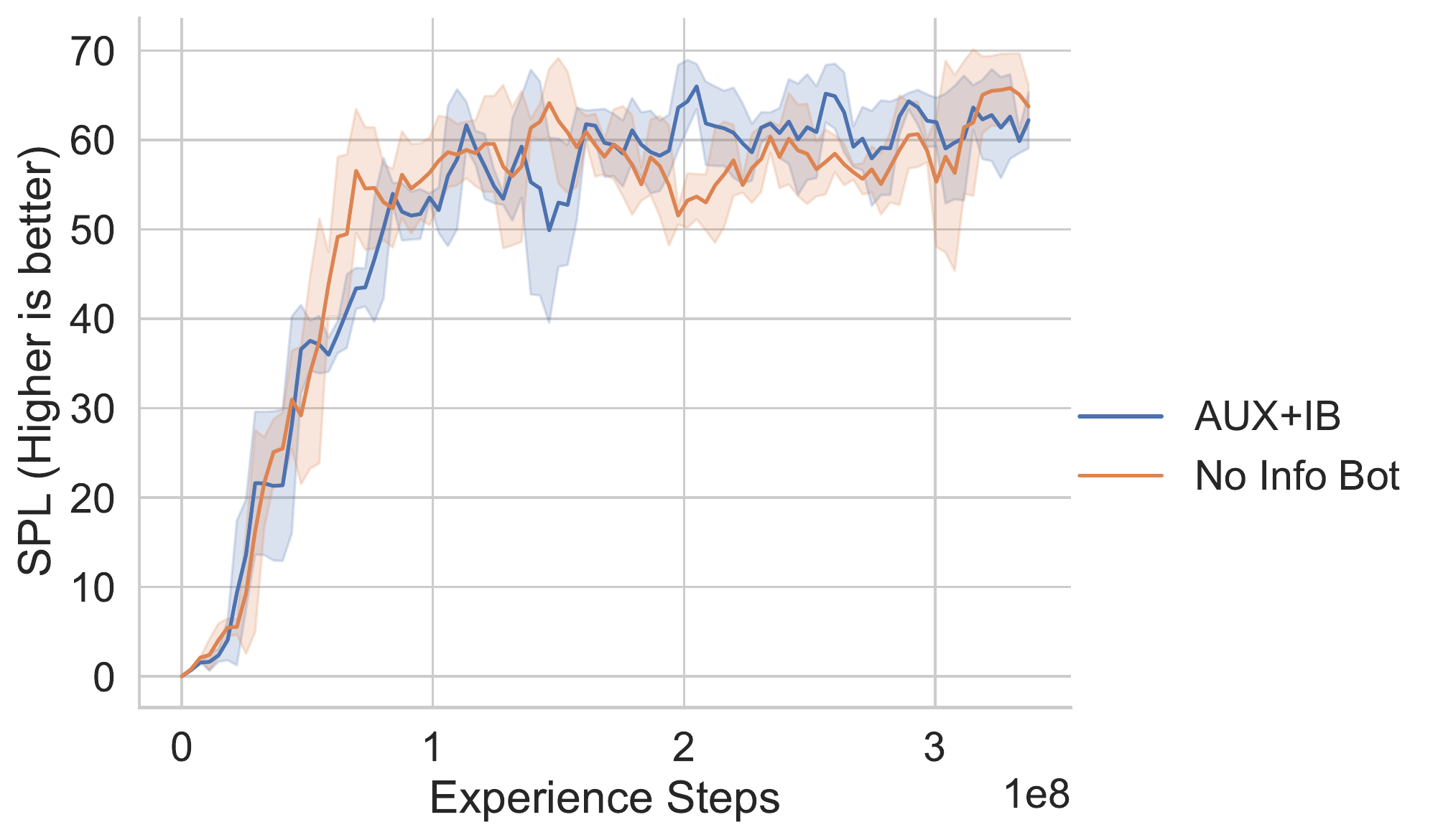}
  \includegraphics[width=0.45\linewidth]{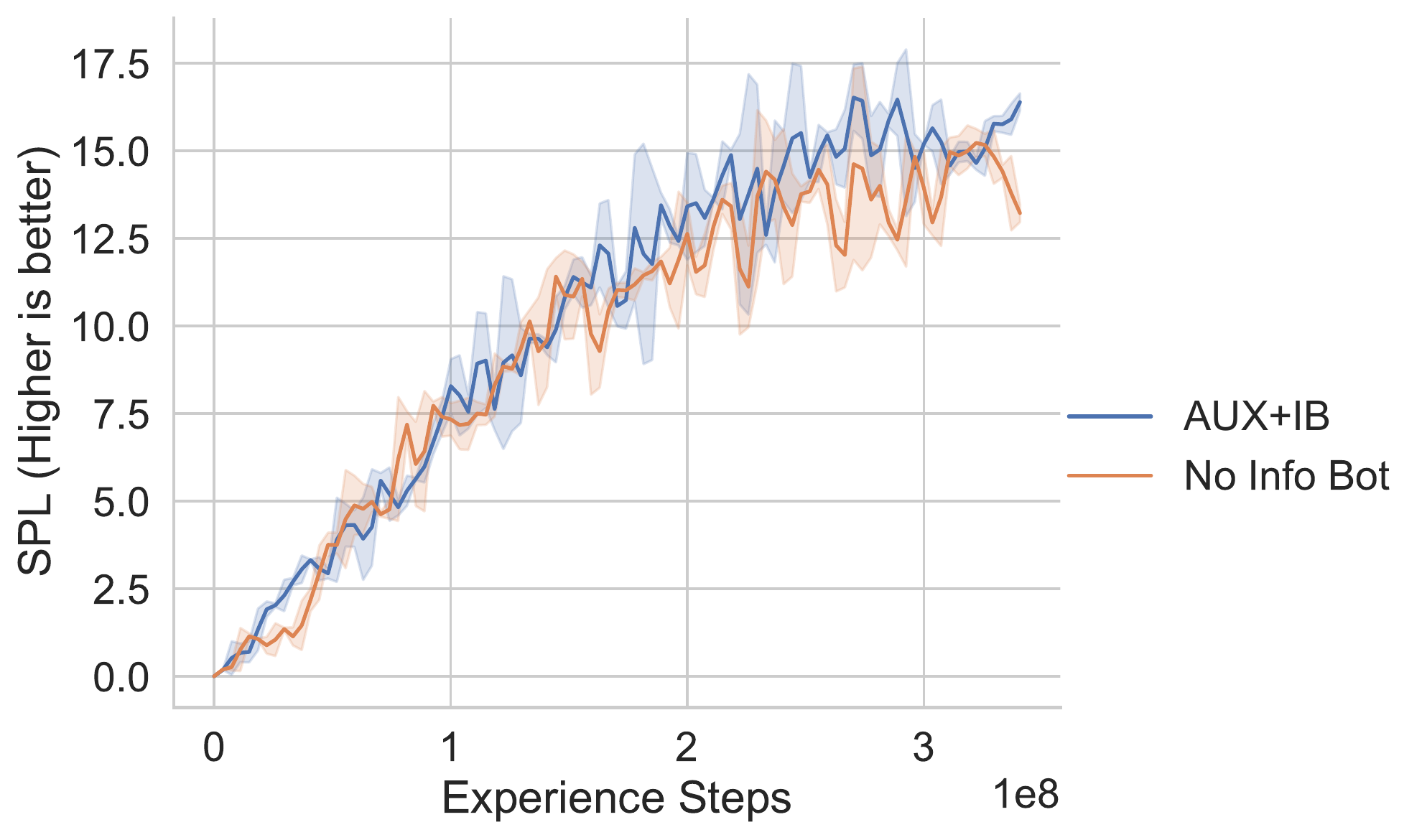}
  \caption{\textbf{Semi-Idealized (left) and Realistic (right) Setting SPL Vs. Steps (DSR)}. While both agent's presented include auxiliary losses to incentives the learning self-localization, only AUX+IB has access to privileged information through an information bottleneck. Including the information bottleneck yields slightly higher navigation performance due to the lower MSE errors in self-localization for both the semi-idealized and realistic settings during training.}\label{fig:nav}
\end{figure*}  

\begin{figure*}
\centering
  \includegraphics[width=0.45\linewidth]{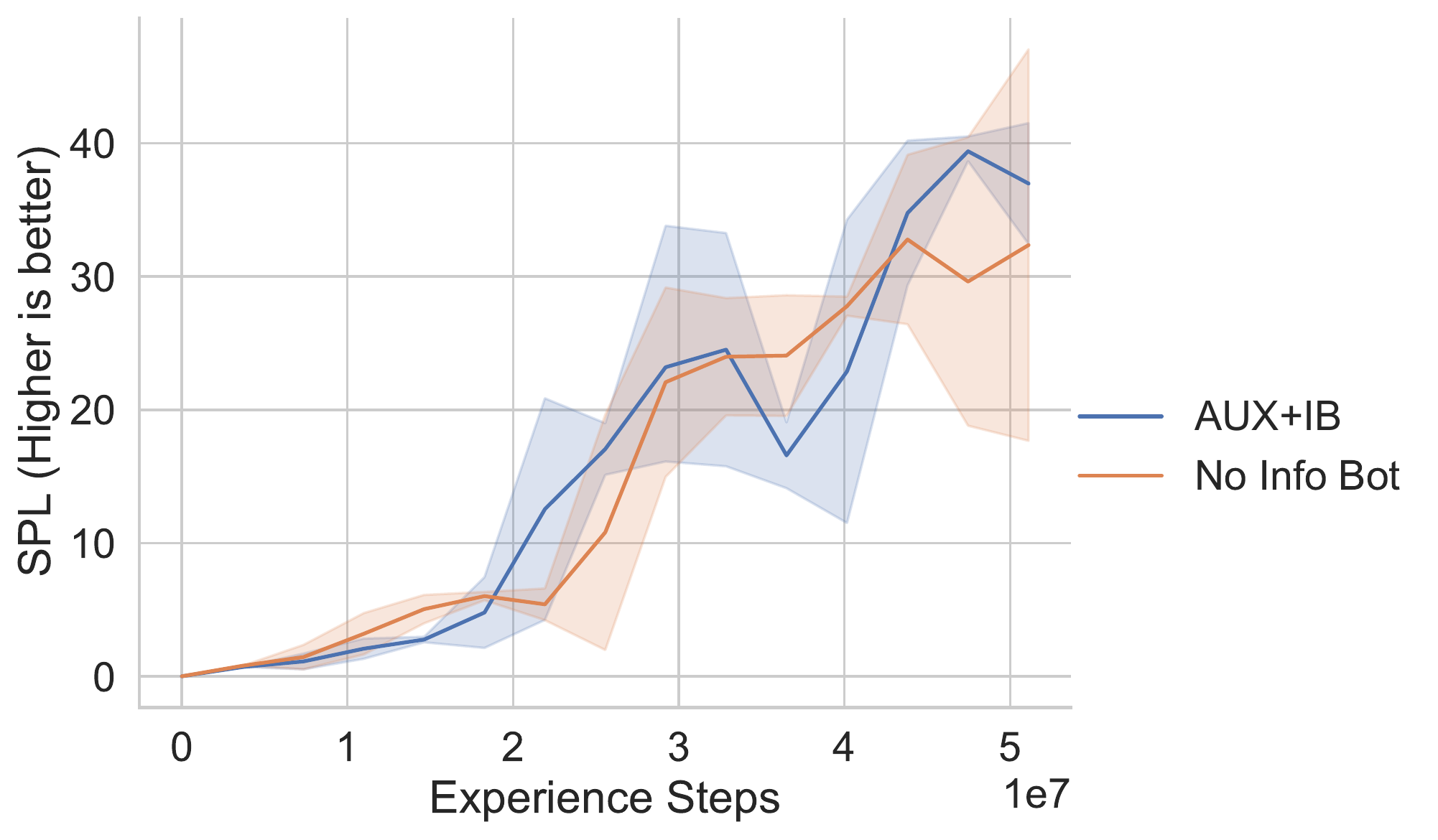}
  \includegraphics[width=0.45\linewidth]{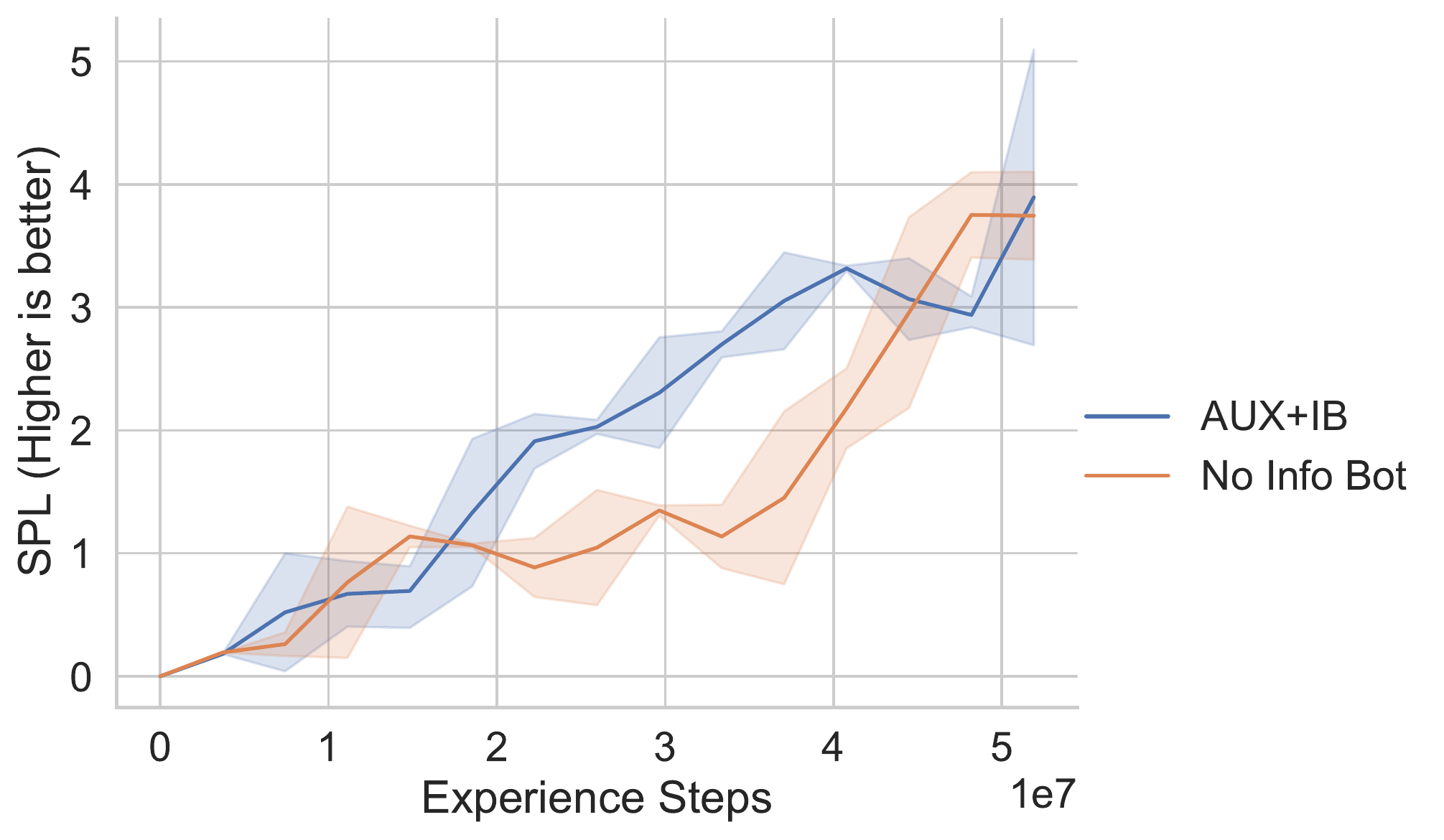}
  \caption{\textbf{Semi-Idealized (left) and Realistic (right) Setting SPL Vs. Steps (DSR, Zoomed In)}. Here we focus on the initial portions of the curves shown in \reffig{fig:nav}. This alternate view of the data exposes the immediate effect that the information bottleneck has on the SPL of the agent, improving sample efficiency. Note, this is much more noticeable in the realistic setting, where the information bottleneck has a larger impact on the overall tracking precision.}\label{fig:nav_zoom}
\end{figure*}  

\begin{figure*}
\centering
  \includegraphics[width=0.45\linewidth]{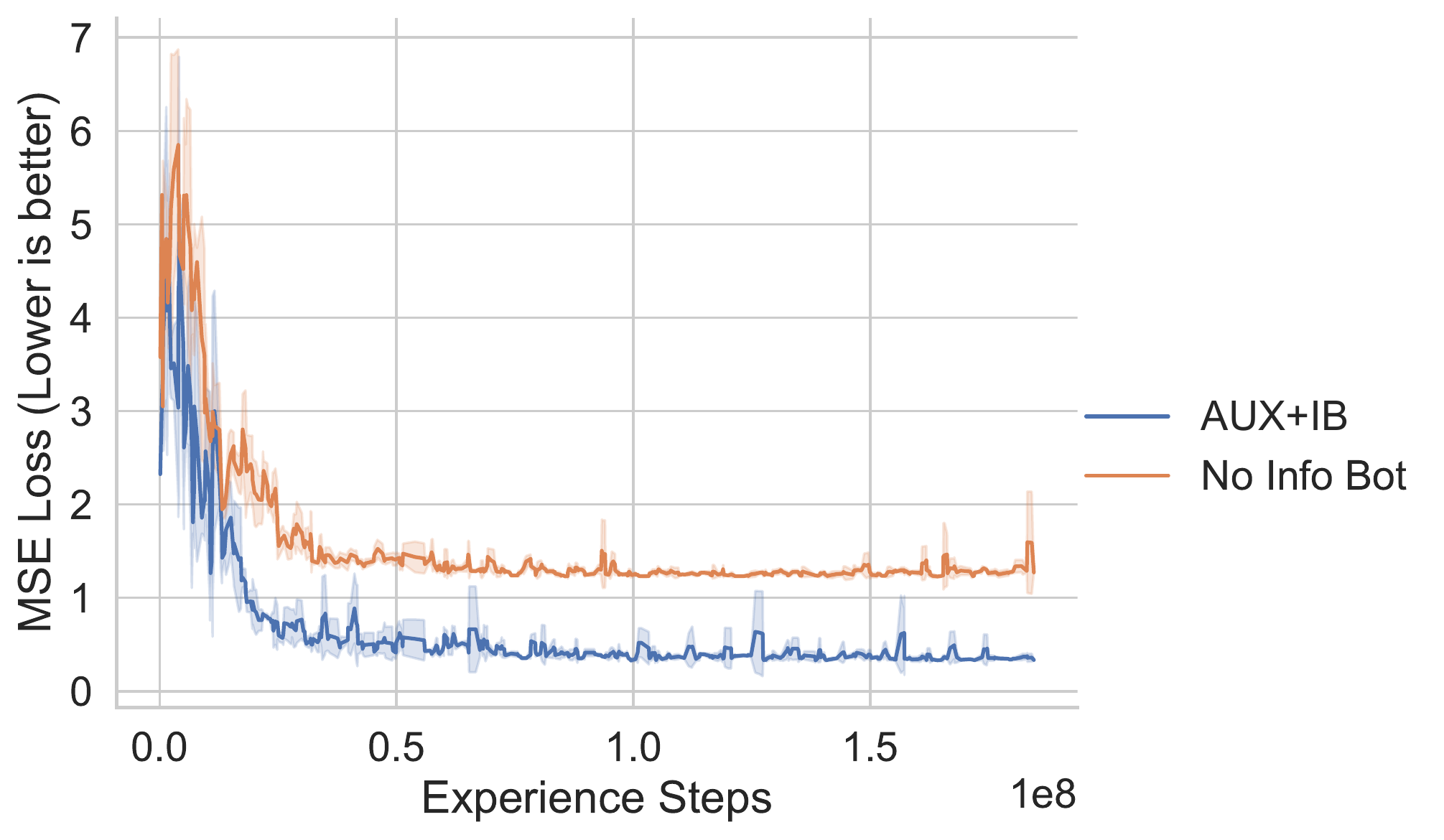}
  \includegraphics[width=0.45\linewidth]{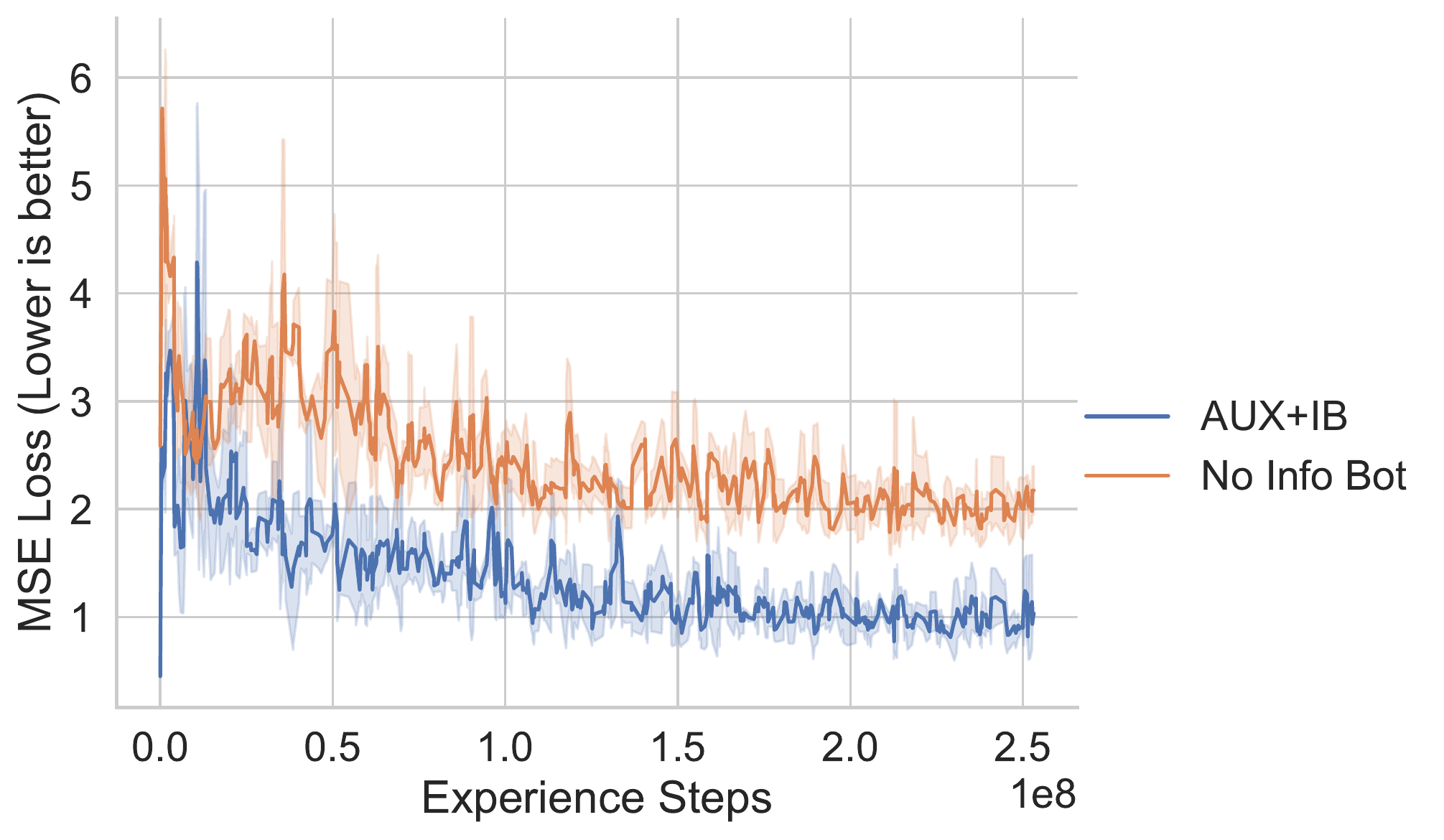}
  \caption{\textbf{Semi-Idealized (left) and Realistic (right) Setting Auxiliary MSE Losses Vs. Steps (DSR)}. While both agent's presented include auxiliary losses to incentives the learning self-localization, only AUX+IB has access to privileged information through an information bottleneck. Including the information bottleneck yields lower MSE errors in self-localization (sum of all Auxiliary Losses reported in Section 4) for both the semi-idealized and realistic settings during training, indicating better tracking performance. As such, we argue that serializing the learning of navigation and self-localization via an information bottleneck yields better performance when jointly learning navigation and tracking in the absence of idealized localization.}\label{fig:aux_losses}
\end{figure*}  

\begin{figure*}
\centering
  \includegraphics[width=0.45\linewidth]{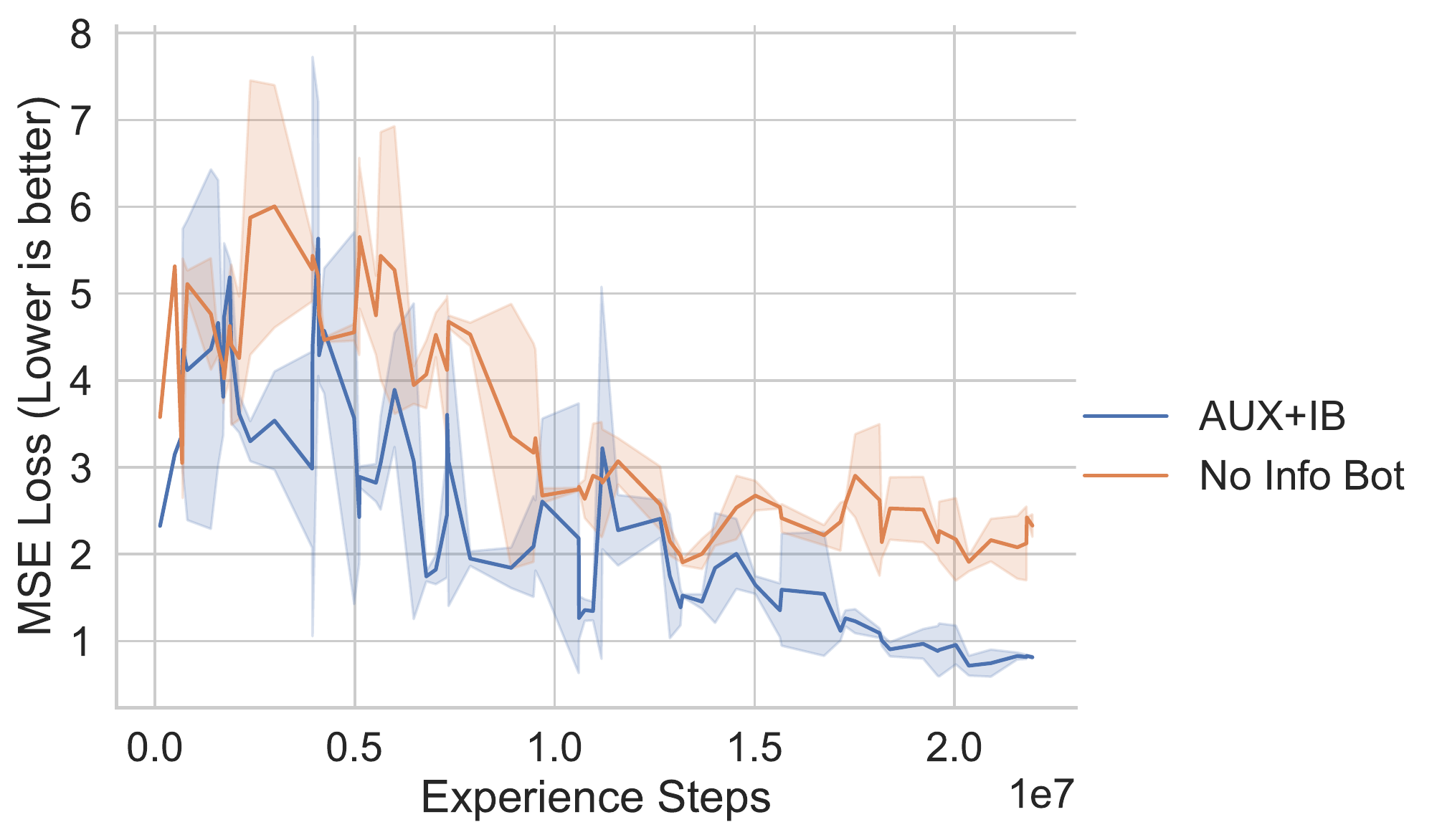}
  \includegraphics[width=0.45\linewidth]{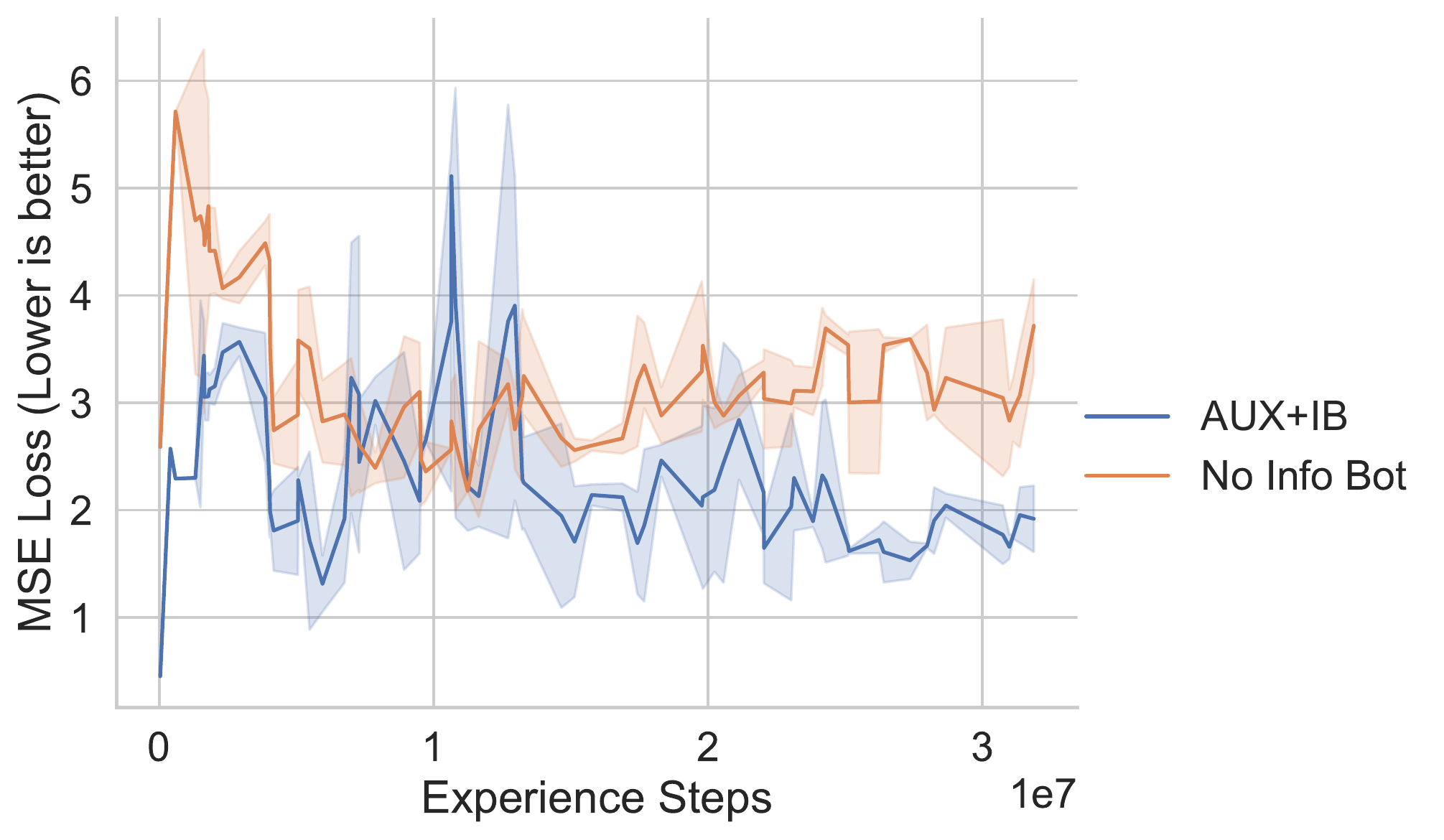}
  \caption{\textbf{Semi-Idealized (left) and Realistic (right) Setting Auxiliary MSE Losses Vs. Steps (DSR, Zoomed In)}. Here we focus on the initial portions of the curves shown in \reffig{fig:aux_losses}, where a divergence between the self-localization MSE losses can begin to be observed. This alternate view of the data exposes spikes in the loss curves both with and without the information bottleneck, illustrating the detrimental effect that changes in the navigation policy has on self-localization. Adding the information bottleneck to serializing the learning of navigation and self-localization yields better self-localization performance even early in training.}\label{fig:aux_losses_zoom}
\end{figure*}  

\begin{figure*}
\centering
  \includegraphics[width=0.90\linewidth]{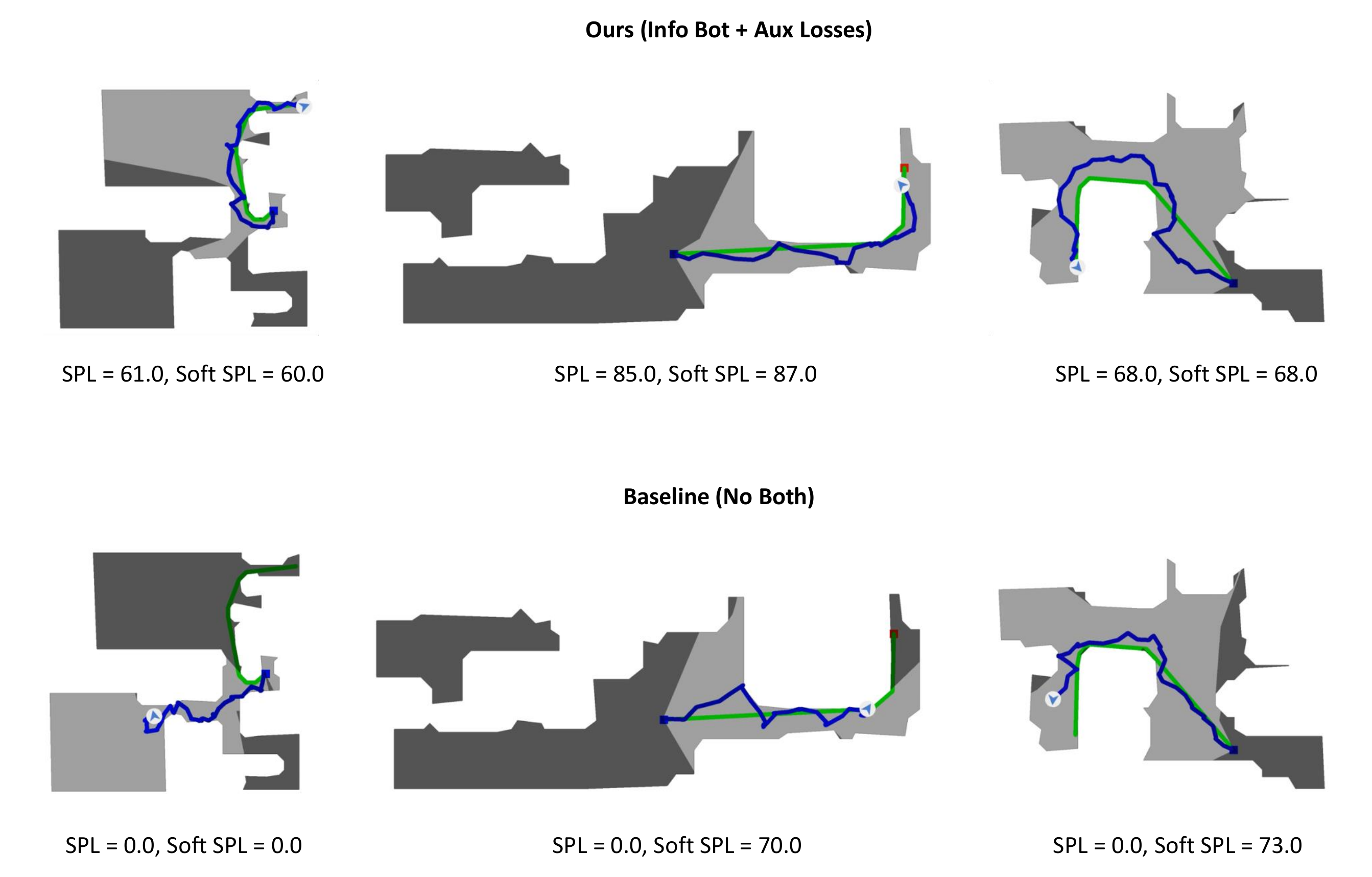}
  \caption{\textbf{Sample Trajectories for Our Agent (above) and Baseline Agent (below)}. The blue curve depicts the agent's path, while the green curve depicts the optimal oracle path. While the absence of auxiliary losses does affect the agent's ability to precisely self-localize, preventing it from stopping within the threshold distance of the PointGoal, it does not hamper the agent's ability to navigate towards the goal. This is evidenced by high Soft-SPL values for both AUX+IB and the No Both baseline in this PointGoal navigation episode.}\label{fig:trajectories}
\end{figure*}  

\begin{figure*}
\centering
  \includegraphics[width=0.45\linewidth]{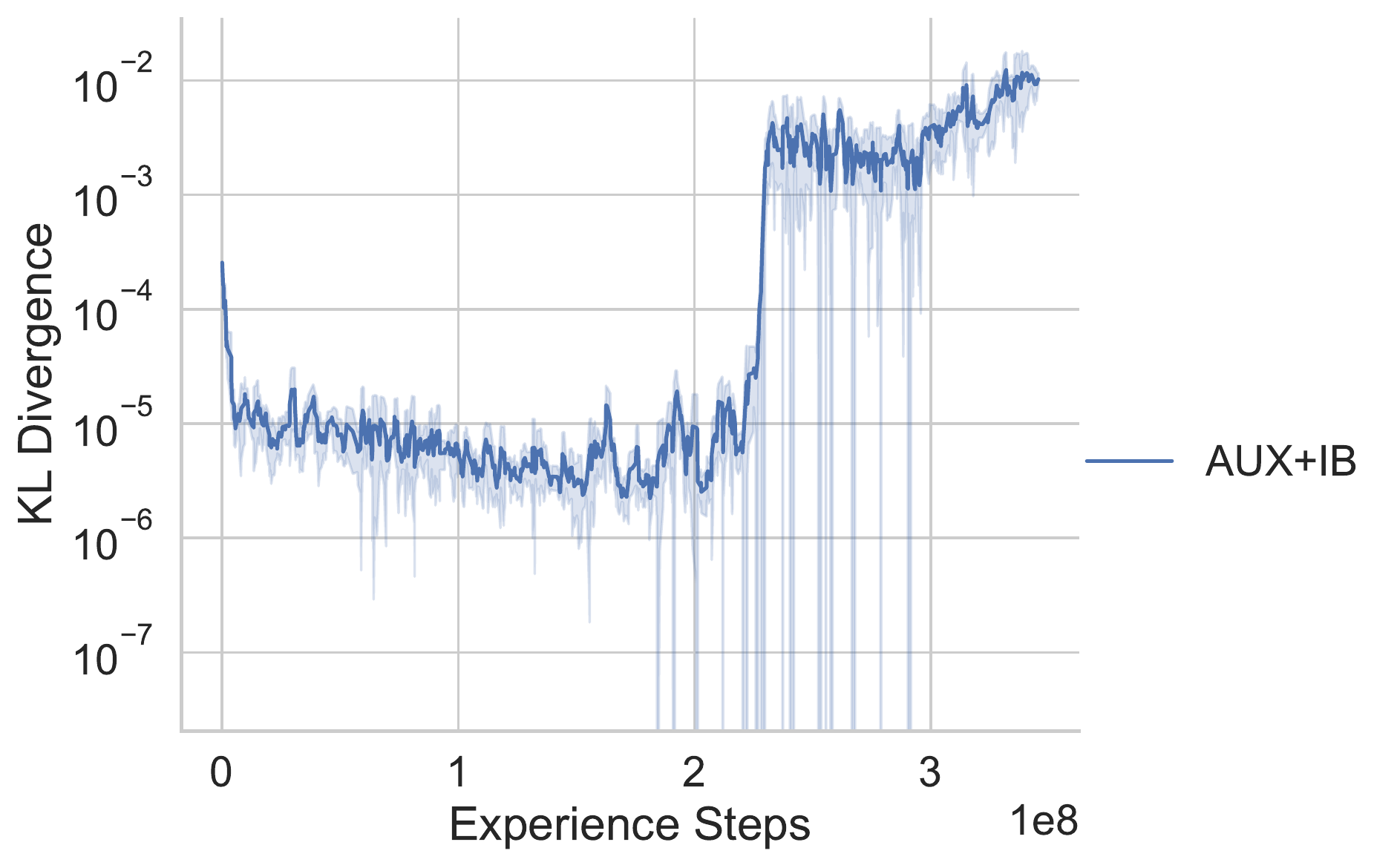}
  \includegraphics[width=0.45\linewidth]{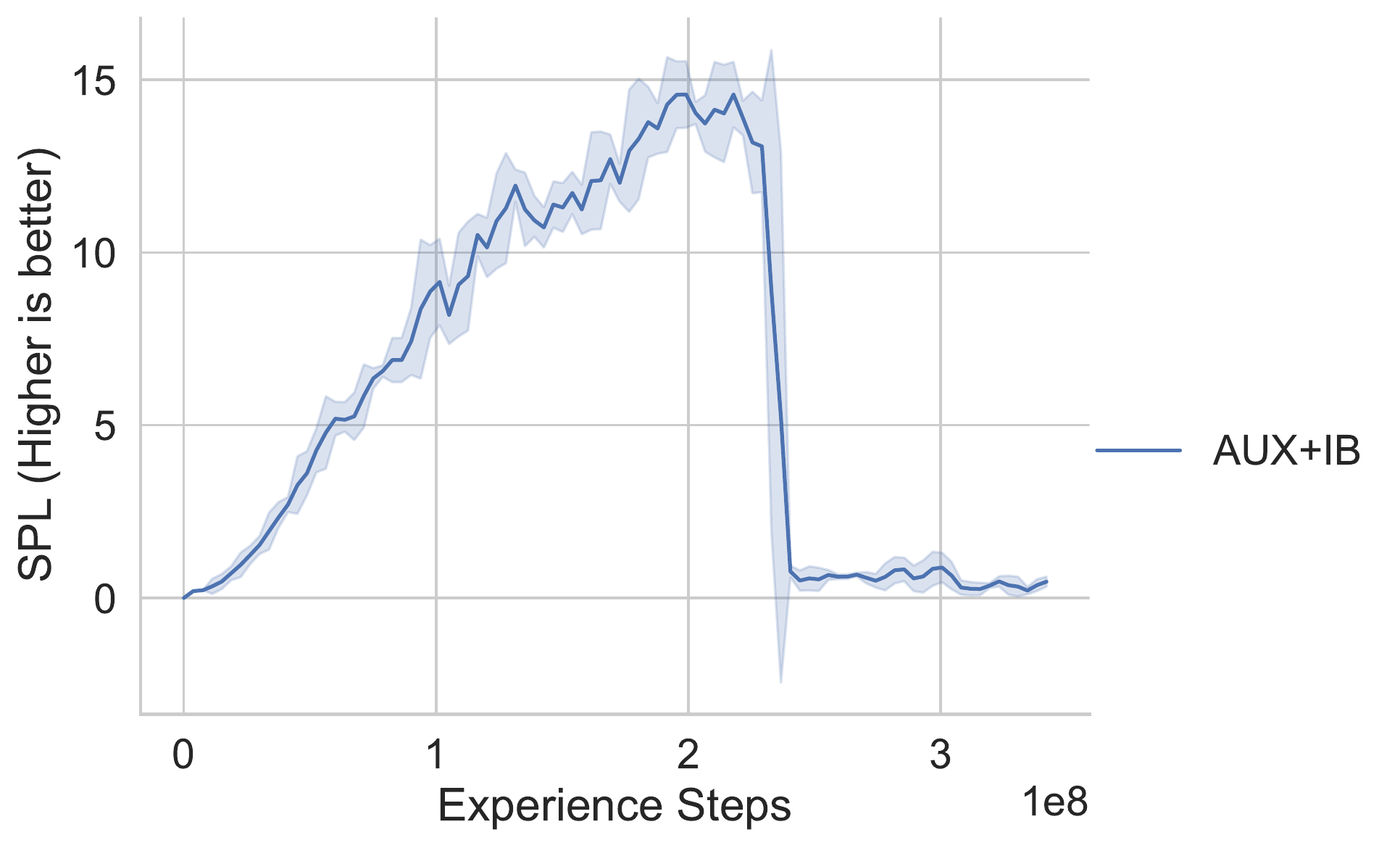}
  \caption{Realistic Setting KL Divergence between $p(Z | S, G)$ and $N(0, I)$ prior (left) and SPL (right) Vs. Steps ($\beta = 1.0$, DSR). Using too small of a fixed beta value promotes the overuse of privileged information at training time. This yields poor performance at evaluation time once the policy learned by the agent becomes reliant on privileged information.}\label{fig:beta_1}
\end{figure*}  

\begin{figure*}
\centering
  \includegraphics[width=0.45\linewidth]{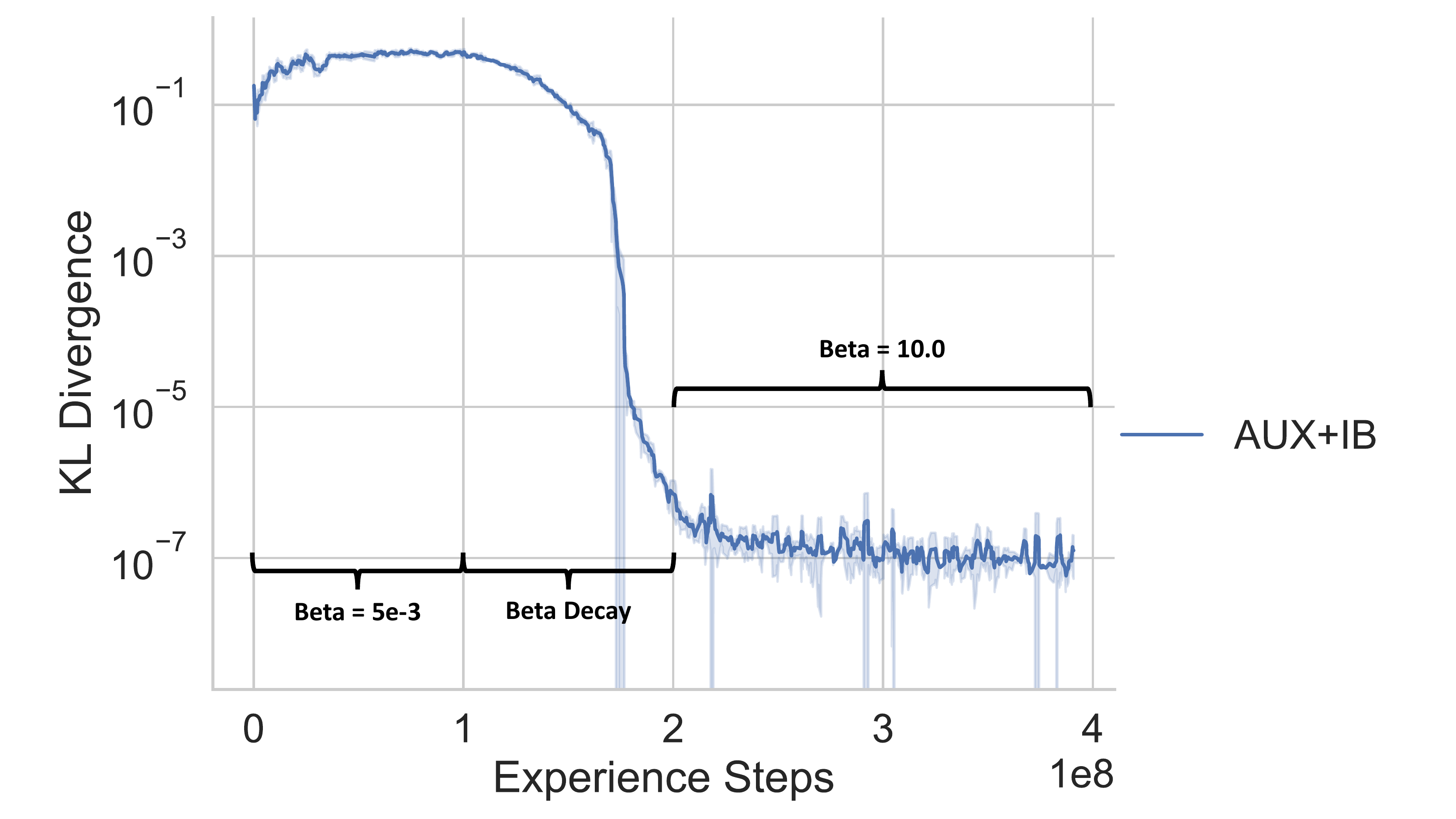}
  \includegraphics[width=0.45\linewidth]{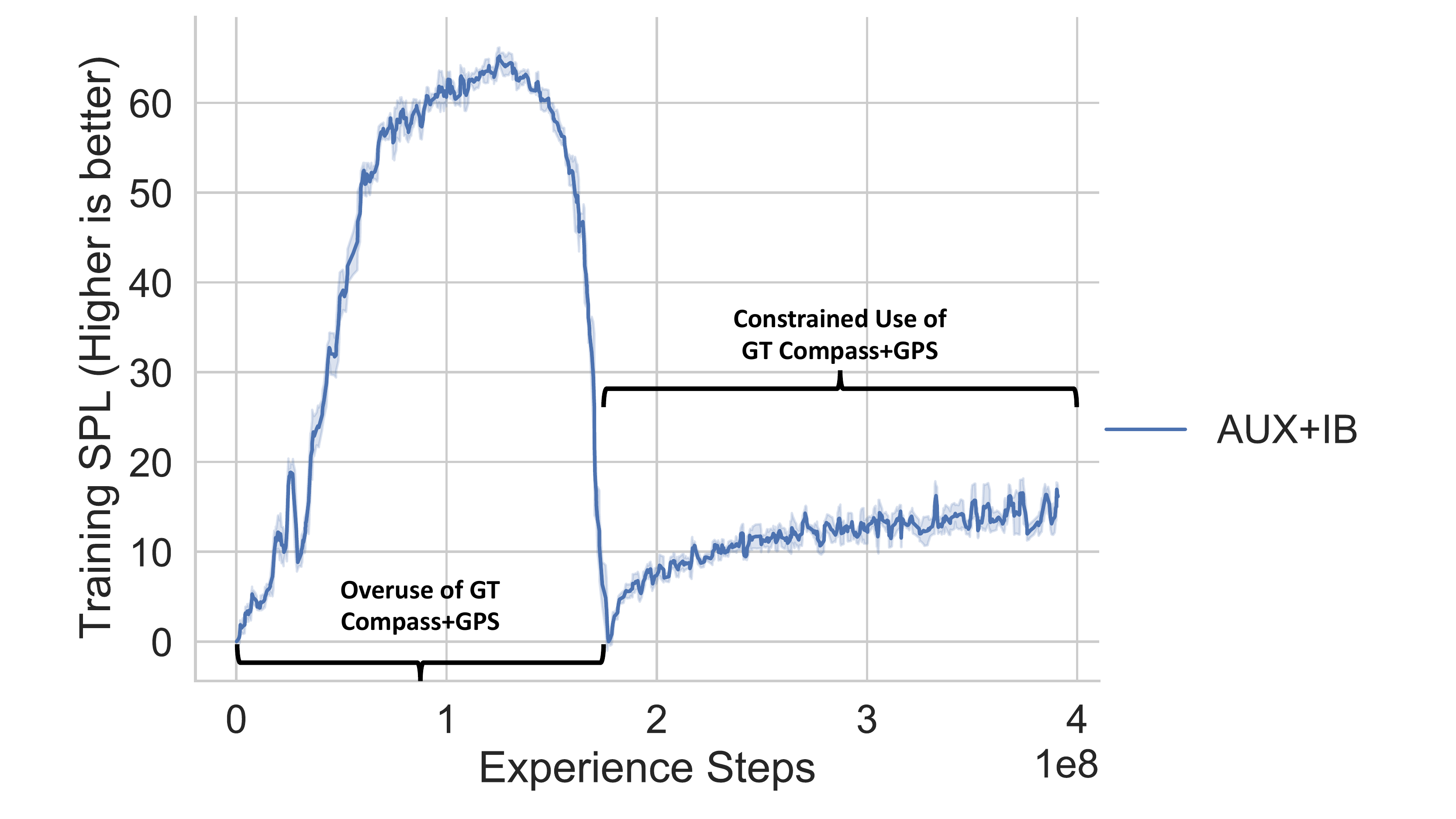}
  \caption{Realistic Setting KL Divergence between $p(Z | S, G)$ and $N(0, I)$ prior (left) and SPL (right) Vs. Steps ($\beta$ decay, DSR). The training curve presented indicates that the agent's policy learned with low beta is unable to adapt to a high beta regime, requiring the agent to relearn a policy from scratch once the final beta region is reached.}\label{fig:beta_5e-3}
\end{figure*}

\end{document}